\documentclass{article}
\usepackage{iclr2027_conference,times}

\usepackage{amsmath,amsfonts,bm}

\def\eqref#1{equation~\ref{#1}}

\def\1{\bm{1}}

\DeclareMathAlphabet{\mathsfit}{\encodingdefault}{\sfdefault}{m}{sl}
\SetMathAlphabet{\mathsfit}{bold}{\encodingdefault}{\sfdefault}{bx}{n}

\newcommand{\xhdr}[1]
{\noindent{\bfseries #1}.}

\usepackage{hyperref}
\usepackage{url}
\usepackage{graphicx}
\usepackage{booktabs}
\usepackage{xcolor}

\usepackage{amsmath}
\usepackage{amssymb}
\usepackage{amsfonts}

\usepackage{multirow}
\usepackage{wrapfig}
\usepackage[bottom]{footmisc}
\usepackage{tcolorbox}
\tcbuselibrary{breakable,listings}
\newtcolorbox{promptbox}{breakable,colback=black!3,colframe=black!45,boxrule=0.4pt,arc=1pt,left=4pt,right=4pt,top=3pt,bottom=3pt,fontupper=\small}
\newtcblisting{promptlisting}{breakable,listing only,colback=black!3,colframe=black!45,boxrule=0.4pt,arc=1pt,left=4pt,right=4pt,top=2pt,bottom=2pt,listing options={basicstyle=\ttfamily\scriptsize,breaklines=true,columns=fullflexible,keepspaces=true}}
\definecolor{linkblue}{RGB}{0,72,150}
\hypersetup{colorlinks=true,linkcolor=linkblue,citecolor=linkblue,urlcolor=linkblue}

\title{Fork-and-Flush: Escaping Idea Basins\\ in Autoresearch Agents}

\author{
Ziyang Cai$^{*}$ \quad Christos Ziakas$^{*}$ \quad Vasilis Kontonis \quad Tim Pearce \\
\textbf{Siddhartha Sen \quad Akshay Krishnamurthy \quad Shivam Garg \quad Dimitris Papailiopoulos} \\[4pt]
Microsoft Research
}

\definecolor{takeawayframe}{HTML}{6AAFA8}
\definecolor{takeawayback}{HTML}{EEF6F5}

\newtcolorbox{takeawaybox}{
  colback=takeawayback,
  colframe=takeawayframe,
  boxrule=0.6pt,
  arc=3mm,
  boxsep=0pt,
  left=8pt, right=8pt, top=6pt, bottom=6pt,
  before skip=8pt, after skip=8pt
}

\newcommand{\takeaway}[2][Takeaway.]{%
  \begin{takeawaybox}\textbf{#1}\ #2\end{takeawaybox}%
}
\iclrfinalcopy
\makeatletter
\def\@maketitle{\vbox{\hsize\textwidth
{\LARGE\sc \@title\par}
\noindent\begin{tabular}[t]{@{}l@{}}\bf\rule{\z@}{24pt}\ignorespaces\@author\end{tabular}%
\vskip 0.3in minus 0.1in}}
\makeatother

\begin{document}

\maketitle
\renewcommand{\headrulewidth}{0pt}

\begin{abstract}
Autoresearch agents tackle open-ended problems by repeatedly proposing candidate solutions, evaluating them, and using feedback to guide subsequent experiments. We show that independent runs of the same agent on the same task often plateau at substantially different scores, with gaps that persist even after considerable additional compute. Embedding their candidate artifacts by functional similarity provides further evidence that trajectories remain in localized regions of the solution space, which we call \emph{idea basins}. To help agents escape these basins, we study a simple periodic intervention, \emph{fork-and-flush}. Our method forks the agent into parallel trajectories, each inheriting the accumulated workspace but starting with a fresh chat context. After running each trajectory for a fixed horizon, the agent continues from the highest-scoring one. Across $13$ long-horizon research and engineering tasks, with individual agent runs lasting up to several days, \emph{fork-and-flush} outperformed the single-run and best-of-$N$ baselines by a relative improvement of $66.0\%$ and $44.4\%$, respectively, on the min--max normalized average score under an equal compute budget.
\end{abstract}

\section{Introduction}
\label{sec:intro}

\begin{figure}[!b]\centering
\includegraphics[width=\linewidth]{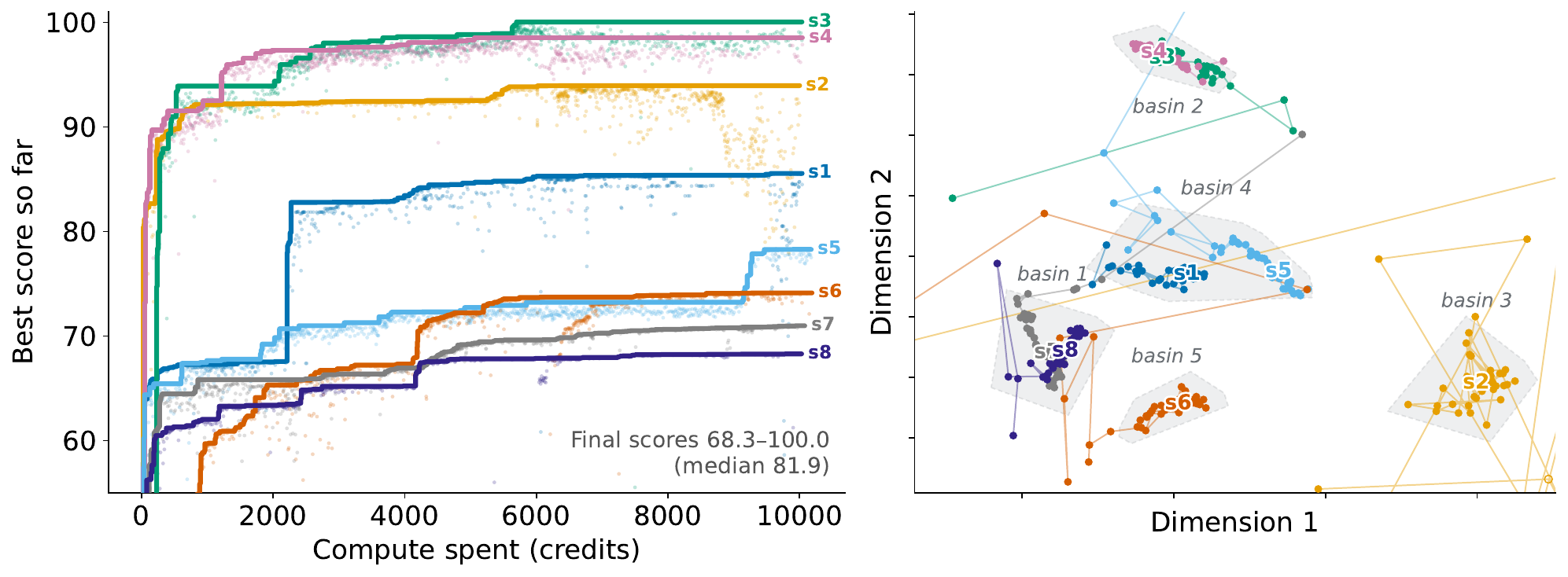}
\caption{Eight independent agents on p171 research task. \textbf{Left:} Best-so-far score against compute spend, with dots representing individual submissions. Runs plateau at different levels, with gaps persisting even after substantial additional compute. \textbf{Right:} The artifacts from these runs embedded by judged functional similarity. Agents converge into five distinct ``idea basins'' and remain there.
}
\label{fig:baseline-traj}
\end{figure}

\begingroup
\renewcommand\thefootnote{*}
\footnotetext{This work was done during an internship at Microsoft Research. \\
{\scriptsize Emails: \texttt{zc5794@princeton.edu}, \texttt{c.ziakas24@imperial.ac.uk},
\texttt{\{shigarg,dimitriosp\}@microsoft.com}}}
\endgroup

Large Language Model (LLM) agents have shown remarkable promise in autonomously solving hard, open-ended research problems requiring long execution horizons \citep{mang2026frontiercs,merrill2026tbench,dillmann2026tbscience}. A prominent paradigm for addressing these problems is the \emph{autoresearch} loop \citep{karpathy2026autoresearch,novikov2025alphaevolve}, where an agent iteratively proposes a solution, evaluates it within an environment, and uses the execution feedback or submission score to guide its next experiment. This process can be viewed loosely as \emph{idea descent}, by analogy with gradient descent: feedback guides successive modifications through the space of candidate solutions. Consider, for example, the task of finding the smallest CIFAR classifier that achieves a target accuracy. Although the objective is easy to state, the design space is broad: the agent can explore different architecture families, layer depths and widths, computational blocks, and training recipes, with many possible tradeoffs between accuracy and model size. A single autoresearch run can evaluate hundreds of candidate solutions and continue for multiple days. This scale raises a fundamental question: over such long horizons, does an agent explore the solution space broadly, or does it mainly refine the ideas it selected early on?

\looseness=-1

We first examine how solution quality evolves over long autoresearch runs. The score of the best solution found so far tends to plateau within a run. More revealingly, independent runs of the same agent on the same task plateau at substantially different levels, with gaps that persist even after considerable additional compute (Figure~\ref{fig:baseline-traj}). This suggests that runs may differ not just in how quickly they improve, but in which parts of the solution space they explore. To investigate this possibility, we embed their candidate artifacts in two dimensions based on functional similarity (Figure~\ref{fig:baseline-traj}). The resulting visualization provides additional evidence: submissions from individual runs tend to remain in localized regions of this space. Continuing the \emph{idea descent} analogy, we call these regions \emph{idea basins}. Some runs appear to spend much of their subsequent effort refining solutions in less productive regions rather than discovering the higher-scoring solutions found by other runs. Notably, this localization persists despite periodic compaction of the agents' conversational histories.

Can we exploit the variation across independent runs to improve an ongoing run by encouraging it to explore more broadly? To this end, we consider two natural operations that are easy to incorporate into existing autoresearch loops. \emph{Forking} lets the agent try several candidate continuations and select one based on their scores. \emph{Flushing} clears the inherited chat history while retaining the accumulated workspace, allowing the continuations to explore with less dependence on earlier reasoning. Our intervention, \emph{fork-and-flush}, combines these operations periodically: it forks the agent into parallel trajectories that each inherit a copy of the current workspace and begin with a fresh chat context, runs each for a fixed horizon, and continues from the highest-scoring trajectory.

\begin{figure}[t]
\centering
\includegraphics[width=\linewidth]{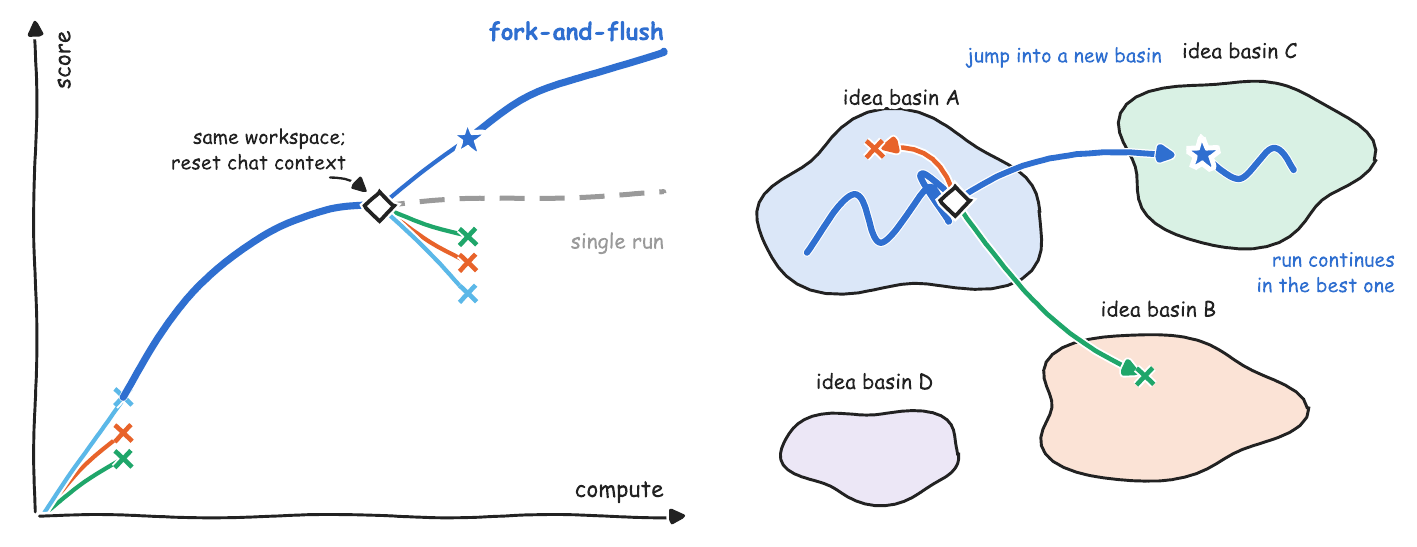}
\vspace{-4pt}
\caption{An illustration of \emph{fork-and-flush}. \textbf{Left:} A single run (grey, dashed) keeps exploring similar ideas within one basin. Fork-and-flush clears the chat context on the same workspace ($\diamond$), fans out $K$ probes, retains the best ($\star$), and discards the rest ($\times$). \textbf{Right:} While the leader's edits remain localized, a fresh-context probe jumps to a new basin via fork-and-flush and continues there. }

\label{fig:teaser}
\end{figure}

We evaluate \emph{fork-and-flush} across 13 long-horizon research and engineering tasks (Table~\ref{tab:tasks}). Eleven are drawn from Frontier-CS \citep{mang2026frontiercs} and EdgeBench \citep{zhu2026edgebench}, spanning problems from polyomino packing and job-shop scheduling to cryptanalysis and inferring a groundwater contaminant source from well readings. The remaining two tasks involve designing a compact CIFAR classifier under an accuracy constraint and optimizing a kernel for a simulated VLIW machine. Individual agent runs last from several hours to multiple days depending on the task. Under an equal compute budget, \emph{fork-and-flush} outperforms both the single-run and best-of-$N$ baselines, achieving $66.0\%$ and $44.4\%$ relative improvements in min--max normalized average score, respectively. Our artifact embedding also shows broader exploration under fork-and-flush (Figure~\ref{fig:teaser}), connecting the gains to the idea-basin observation.

We summarize our key contributions as follows:
\begin{itemize}
\item We show that long-running autoresearch agents exhibit persistent outcome variance: independent runs on the same task plateau at substantially different scores, with gaps that remain even after considerable additional compute.
\item We show that a simple intervention, \emph{fork-and-flush}, improves performance by periodically forking and flushing the chat context while preserving accumulated work. In an evaluation across 13 open-ended tasks, including research and applied engineering challenges with individual runs lasting up to 107 hours, it outperforms single-run and best-of-$N$ baselines under equal compute budgets.
\item We develop an embedding of agent artifacts based on functional similarity to study where trajectories search. It reveals localized exploration consistent with \emph{idea basins} and shows broader movement through this space under \emph{fork-and-flush}.
\end{itemize}

\section{Preliminaries: Autoresearch Setup}
\label{sec:preliminaries}

\xhdr{Open-ended Research Tasks}
We evaluate on thirteen diverse open-ended tasks spanning algorithmic optimization, cryptanalysis, scientific inference, and systems engineering (Table~\ref{tab:tasks}). Nine are drawn from Frontier-CS~\citep{mang2026frontiercs}, including four classical algorithmic challenges (\texttt{polyomino}, \texttt{job-shop}, \texttt{sphere-pack}, \texttt{relay-lay}), three AtCoder competitive programming ports (\texttt{AHC014}, \texttt{p150}, \texttt{p171}), one cache optimization research task (\texttt{LLM-SQL}), and one cryptanalysis challenge (\texttt{LWE-rec}). Two tasks come from EdgeBench~\citep{zhu2026edgebench}, targeting groundwater source recovery (\texttt{Borden-inv}) and black-box permutation optimization (\texttt{order-add}). The remaining two are large-budget engineering problems, specifically Anthropic's public VLIW optimization take-home (\texttt{kernel})~\citep{anthropic2026takehome} and a custom CIFAR model compression task (\texttt{tinyCIFAR}). Performance is evaluated using verifiable, deterministic metrics tailored to each domain, providing an unbounded signal for continuous improvement. Because these raw scores use task-specific scales, measuring outcome variance and comparing interventions requires multiple independent trajectories on each task. Per-run budgets range from $6{,}000$ to $100{,}000$ GitHub Copilot CLI credits, with individual runs lasting from several hours to multiple days. Full task descriptions are provided in Appendix~\ref{app:tasks}.

\xhdr{Autoresearch Harness}
An agent harness enables the interaction between an LLM agent and an isolated container environment~\citep{falck2026faraday}. In our autoresearch setup~\citep{karpathy2026autoresearch}, the agent's execution proceeds as a sequence of submission episodes including editing, execution, and evaluation in a loop; within an episode, the agent edits its solver, runs it, and submits for a score as often as it chooses. These episodes terminate upon a self-halting action by the agent or by reaching a predefined token budget. Upon the conclusion of an episode, our autoresearch harness resumes the agent via a short continuation prompt. Across these episodes, both the \emph{workspace} $W$, consisting of the files it has produced, and the \emph{conversational context} $C$, its accumulated chat history, are strictly preserved, and interventions occur only at episode boundaries. We define a state as $s_i = (W_i, C_i)$ and a \emph{lineage} $\ell = (s_0, \dots, s_t)$ as the trajectory of these states, starting from an empty base state $s_0 = (\emptyset, \emptyset)$. At each step, a deterministic grader evaluates the workspace to yield a score $f(W_i)$, and we let $J(\ell) = \max_i f(W_i)$ denote the maximum objective score achieved by this lineage. For a given task with a fixed total compute budget $B$, measured in compute \emph{credits}, a fixed function of input/output model tokens ($100$ credits $\approx$ US\$1), the final result is the maximum score discovered by an autoresearch agent across a long-horizon run $R = \max_{\ell \in \mathcal{L}} J(\ell)$, where $\mathcal{L}$ represents the set of all lineages instantiated by the method within budget $B$. All conditions use the same coding agent, GitHub Copilot CLI (version 1.0.73), with the same tools, prompt, and task workspace.

\section{Idea Basins in Autoresearch}
\label{sec:basins}

\xhdr{Persistent Variance} We begin by examining independent autoresearch runs on a single task. Figure~\ref{fig:baseline-traj} (left) shows the best score found so far by eight runs on \texttt{p171}. Although the runs share the same initial state and underlying model, they plateau for long stretches at substantially different levels. Their final best scores range from 68.3 to 100.0, and the gaps remain wide even after considerable additional compute. The combination of within-run plateaus and persistent between-run variation suggests that the runs differ not just in how quickly they improve, but also in which parts of the solution space they explore. Some appear to spend much of their remaining compute refining less promising approaches without reaching the higher-scoring approaches found by other runs. In Appendix~\ref{app:variance}, we provide additional plots demonstrating this wide spread of independent runs on four further tasks.

\xhdr{Idea Basins} A submission may be a complex program, and its score tells us little about how it relates to other submissions. Can we visualize these relationships to see what an agent explores as its score plateaus? To do so, we ask an LLM to compare artifacts in triplets: given an anchor artifact $a$ and two candidates $b$ and $c$, which candidate takes an approach more similar to $a$? We then seek a two-dimensional map that respects these relative comparisons. To construct the triplets, we collect graded submissions, remove exact duplicates by content hash, and subsample up to $51$ evenly spaced artifacts per run. Before comparison, we rewrite each artifact as a functional specification under a fixed schema. This reduces cues from code syntax and shared scaffolding that might make submissions from the same run look similar for reasons unrelated to their approach. We fit the map using t-distributed stochastic triplet embedding (t-STE)~\citep{maaten2012tste}.

The map captures substantial structure in the LLM's similarity judgments. Across five mapped tasks, it agrees with $85\%$ of held-out triplet judgments on average, compared with $50\%$ for chance and $56\%$ for a heuristic that knows only which run produced each artifact. Figure~\ref{fig:baseline-traj} (right) shows the map for \texttt{p171}. Submissions from individual runs tend to concentrate in localized regions, which we call \emph{idea basins}. The shaded regions show $k$-means clusters fitted to baseline artifacts. Later submissions generally remain near the same region, providing a view of the repeated refinement suggested by the score plateaus. To test whether this clustered pattern depends on our choice of t-STE, we repeat the mapping using Soft Ordinal Embedding (SOE)~\citep{terada2014soe}, achieving a mean held-out agreement of $0.83$. In addition, the observed clustering of idea basins qualitatively remains the same under SOE, albeit with less sharp boundaries, as described in Appendix~\ref{app:soe}. The full mapping pipeline and the validation of both embeddings are detailed in Appendix~\ref{app:mapping}. In the following sections, we introduce the \emph{fork-and-flush} intervention and use the same map to show how it helps trajectories move between idea basins.

\takeaway{Independent autoresearch runs on the same task often plateau at vastly different performance levels. By clustering artifact embeddings based on their functional similarity, we observe that solutions from individual runs tend to remain in localized regions of this space, which we call idea basins. This motivates the question of whether we can exploit the variation across independent runs to improve an ongoing run by encouraging it to explore more broadly.}

\section{Escaping Idea Basins via Fork-and-flush}
\label{sec:methodology}

\xhdr{Problem Setup}
Open-ended research problems present a combinatorially large solution space, resulting in significant outcome variance across agent trajectories, as empirically shown in Section~\ref{sec:basins}. This raises a natural question: can we leverage the variance observed across independent runs to help an ongoing trajectory escape its current basin? Committing a global budget $B$ entirely to a single run (pure exploitation) often leads the agent to settle into one basin early and spend the remaining budget on marginal refinements. Conversely, parallel scaling approaches like best-of-$N$~\citep{snell2024scaling} represent pure exploration: $N$ independent runs of budget $B/N$ sample diverse basins, but the individual runs remain shallow, and the accumulated workspaces of the $N{-}1$ discarded runs are completely wasted. Interleaved forking provides a natural interpolation between these extremes. A process that never branches is a single run; one that branches exactly once from an empty workspace and terminates is best-of-$N$. To actively exploit this  variance to escape idea basins, we introduce an approach that alternates between parallel sampling and serial execution, driven by two core mechanisms: forking and flushing.

\xhdr{Forking and Flushing}
Our method relies on two underlying mechanisms to escape idea basins. First, \emph{forking} creates $K$ parallel agents $\ell^1, \dots, \ell^K$. Each branch inherits the current lineage and is allocated an exploration budget $b_{\text{explore}}$. Upon exhausting this budget, the highest-scoring branch $\ell^\star = \arg\max_k J(\ell^k)$ is selected, and the remaining branches are discarded. Forking places $K$ probes into the search space simultaneously, allowing the selection mechanism to choose between basins rather than refining within just one. Second, \emph{flushing} retains the accumulated workspace but flushes the conversational context: $(W, C) \to (W, \emptyset)$. A branch that retained $C$ would carry the parent's reasoning that tends to re-enter the same local basin. Starting from $(W, \emptyset)$ restores search diversity while preserving the accumulated workspace.

\xhdr{The Fork-and-flush Algorithm}
The proposed algorithm alternates between exploration and exploitation on a fixed schedule as shown in Figure~\ref{fig:teaser}. It begins with an initial round of forking from empty workspaces: $K$ agents each receive $b_{\text{explore}}$, and the highest-scoring one is selected as the \emph{leader}. The leader then continues alone until it exhausts an exploitative budget $b_{\text{exploit}}$. At that trigger point, the algorithm initiates a fork coupled with a flush: $K$ new parallel agents start from the current workspace $(W_t, \emptyset)$, and each is granted $b_{\text{explore}}$. When the round closes, the best branch is selected as the new leader, the others are pruned, and the run continues sequentially from the new trunk. This loop repeats until the global budget $B$ is exhausted. The run's final result is the best score found by any agent throughout the trajectory. In Section~\ref{sec:exp-ablations}, we conduct ablation studies to isolate the independent contributions of forking and flushing.  We opt for a fixed schedule to trigger \emph{fork-and-flush} rather than dynamic plateau detection; full details on this design choice and its sensitivity are provided in Appendix~\ref{app:schedule_design}.

\begin{table*}[t]
\centering
\footnotesize
\setlength{\tabcolsep}{4pt}
\caption{\small Final performance for \emph{single-run}, \emph{best-of-4}, and \emph{fork-and-flush} (ours) agents evaluated under matched compute budgets. We report the maximum score, mean score, and standard deviation across 5 independent runs (4 for \texttt{tinyCIFAR}), with the highest max and mean per row highlighted in \textbf{bold}. The aggregate row reports the min--max normalized averages.}

\label{tab:main-results}
\begin{tabular}{lrrr|rrr|rrr}
\toprule
& \multicolumn{3}{c|}{\textbf{single-run}} & \multicolumn{3}{c|}{\textbf{best-of-4}} & \multicolumn{3}{c}{\textbf{fork-and-flush (ours)}} \\
\cmidrule(lr){2-4}\cmidrule(lr){5-7}\cmidrule(lr){8-10}
\textbf{task} & max & mean & sd & max & mean & sd & max & mean & sd \\
\midrule
\multicolumn{10}{l}{\emph{Benchmark tasks: 6{,}000--25{,}000 credits per run, hours to several days each}} \\
\texttt{polyomino} & 94.31 & 88.56 & 4.95 & 94.66 & 90.73 & 3.90 & $\mathbf{95.29}$ & $\mathbf{94.41}$ & 0.64 \\
\texttt{job-shop} & 10.36 & 9.66 & 0.87 & 10.27 & 9.91 & 0.24 & $\mathbf{10.92}$ & $\mathbf{10.33}$ & 0.59 \\
\texttt{AHC014} & 39.35 & 23.71 & 9.18 & 25.11 & 20.66 & 3.41 & $\mathbf{105.57}$ & $\mathbf{57.46}$ & 29.35 \\
\texttt{p150} & 130.80 & 123.47 & 4.90 & 133.25 & 123.47 & 5.99 & $\mathbf{135.37}$ & $\mathbf{129.74}$ & 4.43 \\
\texttt{p171} & $\mathbf{100.20}$ & 84.26 & 13.27 & 99.26 & 88.88 & 8.57 & 98.95 & $\mathbf{92.79}$ & 3.39 \\
\texttt{sphere-pack} & 44.70 & 43.87 & 0.83 & 44.64 & 43.76 & 0.99 & $\mathbf{45.43}$ & $\mathbf{44.94}$ & 0.34 \\
\texttt{relay-lay} & 72.73 & 72.59 & 0.12 & 72.77 & 72.74 & 0.04 & $\mathbf{72.82}$ & $\mathbf{72.81}$ & 0.02 \\
\texttt{LWE-rec} & $\mathbf{158.0}$ & $\mathbf{155.3}$ & 2.37 & 146.00 & 141.14 & 10.30 & 153.00 & 144.60 & 9.31 \\
\texttt{LLM-SQL} & $\mathbf{27.05}$ & 26.90 & 0.18 & $\mathbf{27.05}$ & 26.93 & 0.21 & $\mathbf{27.05}$ & $\mathbf{27.04}$ & 0.01 \\
\texttt{order-add} & 30.29 & 24.60 & 4.45 & 27.23 & 26.23 & 1.16 & $\mathbf{66.78}$ & $\mathbf{49.52}$ & 13.60 \\
\texttt{Borden-inv} & 79.55 & 64.62 & 18.12 & 86.28 & 76.45 & 9.47 & $\mathbf{93.90}$ & $\mathbf{85.85}$ & 5.54 \\
\midrule
\multicolumn{10}{l}{\emph{Real-world tasks: 31{,}000--100{,}000 credits per run, several days each} } \\
\texttt{kernel} & 132.96 & 120.36 & 9.75 & 127.36 & 122.50 & 5.76 & $\mathbf{142.46}$ & $\mathbf{135.23}$ & 4.45 \\
\texttt{tinyCIFAR} & 0.586 & 0.429 & 0.129 & 0.633 & 0.560 & 0.080 & $\mathbf{0.682}$ & $\mathbf{0.572}$ & 0.115 \\
\midrule
\emph{min-max normalized avg.} & 0.75 & 0.47 & 0.26 & 0.73 & 0.54 & 0.21 & $\mathbf{0.98}$ & $\mathbf{0.78}$ & 0.16 \\
\bottomrule
\end{tabular}
\end{table*}

\section{Experiments}
\label{sec:experiments}
\looseness=-1
\subsection{Experimental Setup}
\label{sec:exp-setup}

\xhdr{Baselines}
We compare our method against two strategies under strictly matched global budgets. The \emph{single-run} baseline spends the full budget $B$ sequentially on a single lineage; we execute eight independent single-run baselines per task. The \emph{best-of-$N$} baseline ($N{=}4$) distributes the budget equally, allocating $B/4$ to each of four independent runs and keeping the best final result.

\xhdr{Evaluation Setup}
Each task is allocated a fixed per-run compute budget $B$, as detailed in Table~\ref{tab:tasks}. Within a given task, all conditions utilize the identical underlying agent model: GPT-5.6-terra at high reasoning effort for eleven tasks, and GPT-5.5 at extra-high reasoning effort for \texttt{kernel} and \texttt{tinyCIFAR}. Each task is evaluated across five independent fork-and-flush runs. We report the maximum, mean, and standard deviation of final scores for each task, alongside a min--max normalized average across all thirteen tasks.

\xhdr{Fork-and-Flush Configuration}
Every \emph{fork-and-flush} run in Table~\ref{tab:main-results} adheres to a fixed forced cadence featuring three fan-outs, removing the need for fragile plateau detection. The run opens with an initial round of $K{=}4$ agents launched from empty workspaces, selecting the highest-scoring candidate as the leader. Two subsequent fork-and-flush rounds are triggered strictly by budget consumption---at $33\%$ and $66\%$ of $B$. In these rounds, $K{=}4$ new branches fork from the leader's accumulated workspace but begin with empty chat contexts. Each branch receives a fixed exploration budget $b_{\text{explore}}$ of approximately $0.06B$; the best-performing branch is adopted as the new leader, the others are discarded, and the leader continues serially until the next trigger. We provide a detailed analysis of this schedule design and its hyperparameter sensitivity in Appendix~\ref{app:schedule_design}.

\subsection{Main Results}
\label{sec:exp-main}

Table~\ref{tab:main-results} compares final performance under matched compute budgets. \emph{Fork-and-flush} achieves a min--max normalized average score of $0.78$, compared with $0.47$ for \emph{single-run} and $0.54$ for \emph{best-of-4}, representing relative improvements of $66.0\%$ and $44.4\%$, respectively. These gains extend across the task suite: \emph{fork-and-flush} achieves the highest mean score on 12 of the 13 tasks. Across tasks, its average normalized \emph{mean} score ($0.78$) exceeds even the average normalized \emph{maximum} score of the single-run baseline ($0.75$). Alongside these performance gains, \emph{fork-and-flush} produces more consistent outcomes on average. The average normalized standard deviation decreases to $0.16$, compared with $0.26$ for \emph{single-run} and $0.21$ for \emph{best-of-4}. Relative to \emph{single-run}, the standard deviation decreases on 10 of the 13 tasks. Variability increases on some tasks, including \texttt{AHC014} and \texttt{order-add}, where these increases accompany substantial gains in mean score. Their larger raw standard deviations should therefore be interpreted alongside the higher score levels reached. Overall, \emph{fork-and-flush} combines substantial improvements in average performance with greater consistency across independent runs.

\begin{wraptable}{r}{0.45\linewidth}
    \centering
    \vspace{-10pt}
    \caption{\small Impact of forking and flushing: min--max normalized mean score over 11 tasks. Best in \textbf{bold}, second best in \emph{italics}.}
    \label{tab:ablation}
    \small
    \setlength{\tabcolsep}{4pt}
    \begin{tabular}{lcc|c}
    \toprule
    \textbf{Method} & \textbf{Flush} & \textbf{Fork} & \textbf{Norm.\ avg} \\
    \midrule
    Single-run      & $\times$   & $\times$   & $0.45$ \\
    Flushing      & \checkmark & $\times$   & $0.55$ \\
    Forking     & $\times$   & \checkmark & $\mathit{0.57}$ \\
    \midrule
    \textbf{Fork-and-flush} & \checkmark & \checkmark & $\mathbf{0.71}$ \\
    \bottomrule
    \end{tabular}
    \vspace{-10pt}
\end{wraptable}

\subsection{Ablations: Impact of Forking and Flushing}
\label{sec:exp-ablations}

To isolate the contributions of our core mechanisms, we evaluate the impact of forking without flushing the context, and flushing the context without forking across 11 tasks (excluding the two longest-horizon tasks that are expensive to run). In Table~\ref{tab:ablation}, we report the performance of all four methods as the min--max normalized final score at the task's budget, averaged across all tasks. The flush-only and fork-only ablations achieve scores of 0.55 and 0.57 respectively, compared to 0.45 for the standard single-run baseline, and 0.71 for the full \emph{fork-and-flush} method. This demonstrates that while both individual components outperform the single-run baseline, neither mechanism alone accounts for the full effect. In addition, forking provides a slightly larger improvement than flushing. The complete per-task breakdown is provided in Appendix~\ref{app:ablation}.

\takeaway{\emph{Fork-and-flush} balances exploration and exploitation by periodically forking from the accumulated workspace with fresh chat contexts and continuing from the highest-scoring branch. Across $13$ tasks under matched compute budgets, it improves min--max normalized average score by $66.0\%$ over \emph{single-run} and $44.4\%$ over \emph{best-of-4}.}

\section{Analysis}
\label{sec:analysis}

\subsection{Visualizing Escaping Idea Basins}
\label{sec:analysis-geometric}

\looseness=-1

\begin{figure}[!ht]\centering
\includegraphics[width=\linewidth]{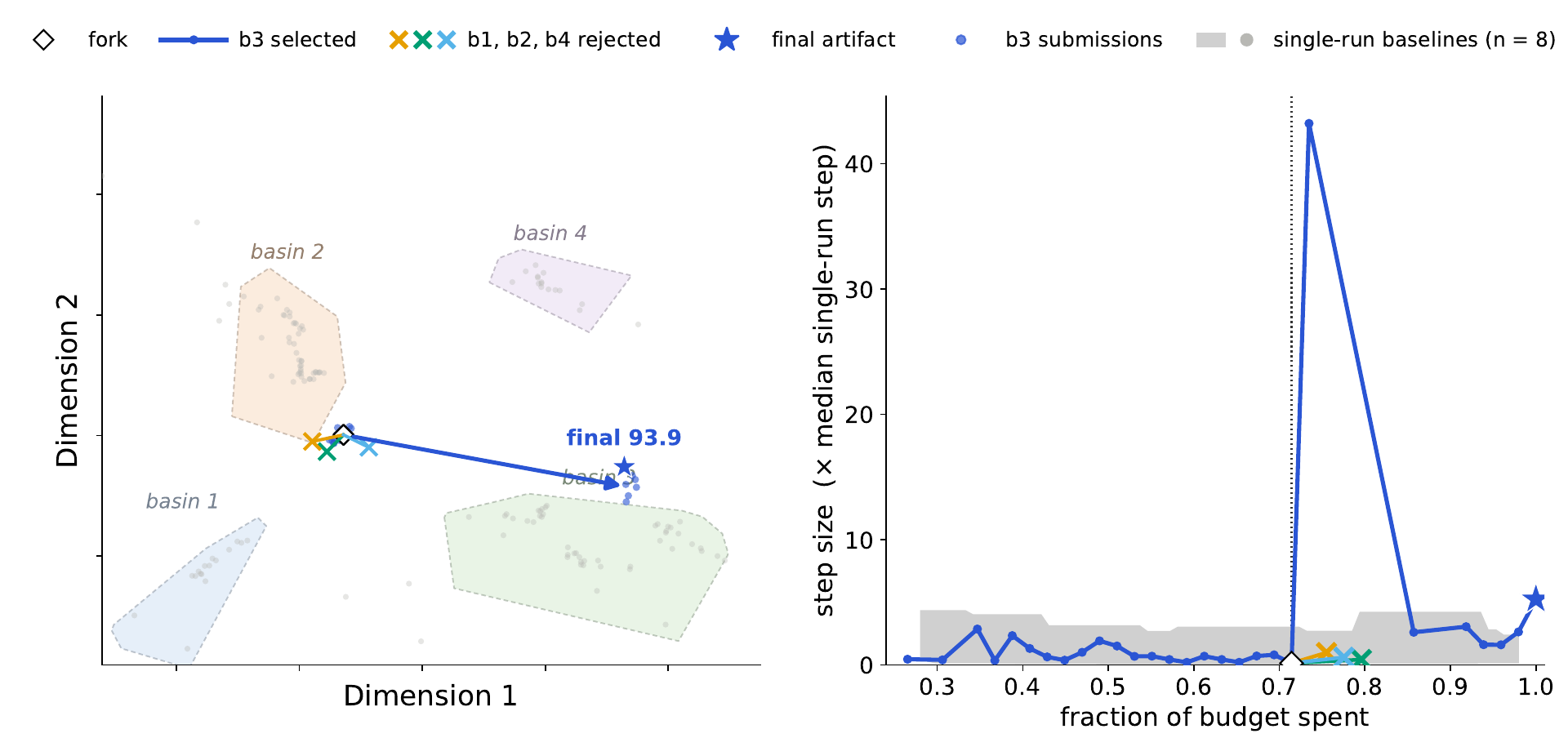}
\caption{\small Fork-and-flush escaping an idea basin on \texttt{polyomino}. \textbf{Left:} A mid-run fan-out overlaid on baseline basins. Rejected branches ($\times$) from the fork point ($\diamond$) remain trapped, but the selected branch (blue arrow) escapes the idea basin to achieve its final score ($\star$). \textbf{Right:} Step distances between consecutive submissions. The selected branch's jump ($\star$) cleanly breaks out of the standard single-run refinement range (grey band).}
\label{fig:basin-gallery}
\end{figure}

\xhdr{Visualizing the Jump}
We place the artifacts of fork-and-flush runs on the same geometric maps built in Section~\ref{sec:basins}, embedded jointly with the single-run baselines so that both share an embedding space, enabling  fork-and-flush to be interpreted as a spatial trajectory across the solution landscape. As illustrated in Figure~\ref{fig:basin-gallery} (left), we overlay the mid-run fan-out of a \emph{fork-and-flush} trajectory on the \texttt{polyomino} task. While leader edits remain highly localized, the intervention successfully forces exploration: branches $b_1$, $b_2$, and $b_4$ remain trapped in the basin of the branch point, but the fresh-context probe $b_3$ successfully jumps into a new region (Basin~3). This probe is selected by the scoring mechanism, allowing the run to escape the local optimum and finish at $93.9$. We provide additional plots showing these jumps in Appendix~\ref{app:jumps}.

\xhdr{Quantifying the Jump} To quantify the jump, we report distances in units of the \emph{median single-run step}, defined as the median displacement between consecutive submissions across all single-run baselines. On \texttt{polyomino}, after the first quarter of the budget, no single run moves more than $4.4$ units between consecutive submissions, and the fork-and-flush leader's own sequential steps stay within $2.9$ units. At the flush fan-out (triggered at $71\%$ of the budget), the winning branch ($b_3$) moves $43$ units in its very first submission after the flush, definitively landing in a different basin; the three rejected branches each move at most one unit and stay where they forked. Once established in the new basin, the leader's refinement steps drop back to the single-run scale (at most $5.2$ units, an order of magnitude below the jump).

\xhdr{Robustness of the Jump} We provide an additional analysis using SOE to fit the embedding space in Appendix~\ref{app:soe}. In our analysis, the qualitative pattern remains identical, confirming the jump is a genuine artifact of our search strategy and robust to the triplet embedding method. In particular, under SOE, no single run moves more than $4.1$ units between consecutive submissions. At the flush fan-out, the three rejected branches each move under one unit, while the winning branch moves $10.1$ units in its first submission after the flush into a different cluster; its subsequent steps then fall back inside the single-run range.

\takeaway{Embedding the artifacts of baseline and fork-and-flush runs jointly, by judged functional similarity, makes the mechanism visible. Single runs each stay inside one idea basin for their whole budget. Fork-and-flush runs jump between basins and explore several: the context-cleared probes land in different basins and selection carries the run into a new one, which on polyomino ends at $93.9$ against a baseline median of $84.6$.}

\subsection{What Drives Variance?}
\label{sec:analysis-variance}

\xhdr{Variance at the capability frontier}
Our hypothesis is that outcome variance is driven by task difficulty relative to the agent's capacity. When a task sits far below the agent's ability, every run finds the same optimal solution and the spread collapses. Conversely, when a task sits far above its capacity, every run fails in the same way, collapsing the spread again. It is only in the intermediate regime---at the exact boundary of what the agent can achieve---that runs make partial, divergent progress, maximizing variance. A natural assumption is that as foundational models scale and become stronger, they will solve current tasks reliably and this variance will vanish. However, we expect that as models scale, the frontier of relevant research tasks will simply scale with them. To empirically test this core hypothesis---that variance peaks at the boundary of capability---we require a domain where difficulty can be dialed continuously, evaluated cheaply, and repeated enough to estimate spread with statistical confidence. Since our thirteen long-horizon research tasks do not permit this, we use MAX-3-SAT on random 3-CNF formulas: the number of variables sets the difficulty, random assignment and a strong classical solver bracket the attainable range, and freshly generated instances successfully rule out memorization.

\begin{wrapfigure}{r}{0.5\linewidth}
   \centering
  \vspace{-8pt}
   \includegraphics[width=\linewidth]{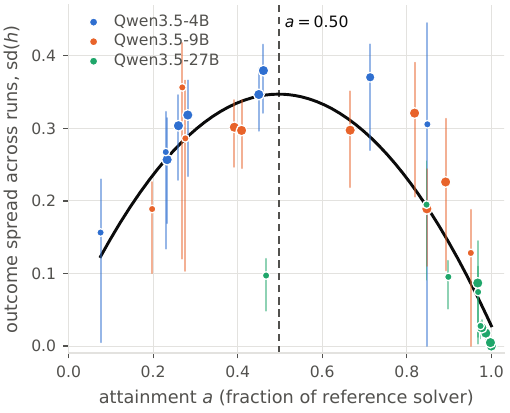}
   \caption{\small Outcome variance is largest at the frontier of what the model can do. Spread
vanishes where every run solves or fails the formula and peaks in between.}
  \label{fig:variance-frontier}
   \vspace{-10pt}
\end{wrapfigure}
\xhdr{Empirical evidence from MaxSAT problem}
We run three open-weight models (Qwen3.5-4B, 9B, and 27B) on random 3-CNF formulas with $V$ variables, $V$ from 10 to $100{,}000$, and measure the number of unsatisfied clauses $u$ of the best assignment each run produces. We normalize this raw quantity into an attainment metric $h$, representing the fraction of the performance gap closed between random guessing and a strong classical solver. Specifically, $h = (u_{\text{random}} - u_{\text{agent}}) / (u_{\text{random}} - u_{\text{ref}})$, where $u_{\text{random}} \approx 6V/8$ is the expected number of unsatisfied clauses for a uniformly random assignment, and $u_{\text{ref}}$ is the minimum $u$ found by a multi-restart WalkSAT baseline. Thus, $h=0$ indicates performance equivalent to random guessing, while $h=1$ indicates matching the WalkSAT reference ceiling. We aggregate independent runs into evaluation cells defined by a specific (model, $V$) pair, with each cell containing between 8 and 32 runs that differ only in their initial seed (32 at most sizes, 8 at the smallest and largest formulas). For each cell, we compute the mean attainment $a$, and the execution spread $s$, defined as the population standard deviation. To quantify the uncertainty in $s$ arising from sample size, we compute $95\%$ confidence intervals via bootstrapping, resampling the runs within each cell with replacement $2{,}000$ times. As shown in Figure~\ref{fig:variance-frontier}, each point represents a single (model, problem size) pair, plotted according to the fraction of the reference solver's performance the model attains (horizontal axis) and the variance among its runs (vertical axis). Because difficulty is measured relative to model capability rather than absolute problem size, the three models trace a unified curve despite operating at different scales. Variance peaks when a model achieves roughly half of the reference performance, collapsing at both the trivial and intractable extremes. Fitting a least-squares quadratic reveals a strict inverted-U relationship between attainment and spread, peaking at $a = 0.497$. This suggests that the observed variance stems from genuine exploration of distinct idea basins, providing the empirical signal that our \emph{fork-and-flush} algorithm exploits.

\takeaway{Outcome variance is highest when a task sits at the frontier of the agent's ability. On a difficulty ladder (MAX-3-SAT), runs agree when the task is trivial or hopeless and disagree most in between, peaking at half the reference solver's attainment. Because open-ended research inherently targets this frontier, we expect that the variance that fork-and-flush exploits will not vanish as models scale.}

\section{Related Work}
\label{sec:related}
\looseness =-1
\xhdr{Autoresearch and Self-Improving Agents}
Long-horizon agents~\citep{wijk2024rebench,chan2024mlebench} are increasingly used to drive open-ended research, from automated paper writing~\citep{lu2024aiscientist,schmidgall2025agentlab} to evolutionary program search~\citep{romera2024funsearch,novikov2025alphaevolve,lange2025robustkbench} and autonomous experimentation~\citep{boiko2023coscientist,swanson2025virtuallab,gottweis2025coscientist,huang2025biomni}. In fact, the research loop proposed by The AI Scientist~\citep{lu2024aiscientist,yamada2025aiscientistv2} automates the idea--experiment--writeup cycle, while AlphaEvolve~\citep{novikov2025alphaevolve} evolves programs against a grader over time. Beyond a single run, LLM agents typically improve \emph{across} tasks either by accumulating experience~\citep{shinn2023reflexion,wang2023voyager,zhou2025memento} or by distilling their own successful trajectories back into the model parameters~\citep{zelikman2022star,singh2024restem,qu2024rise,falck2026faraday}. \emph{Fork-and-flush} is orthogonal to both paradigms as it improves a single run \emph{within} one task, with no memory carried across tasks and no weight updates. The only state that propagates across branches is the workspace, as the intervention exists precisely to clear the conversational context. To evaluate LLM agents in these long-horizon settings, benchmarks such as Terminal-Bench~\citep{merrill2026tbench} and Terminal-Bench-Science~\citep{dillmann2026tbscience} grade end-to-end terminal workflows on a strict pass/fail basis, while Frontier-CS~\citep{mang2026frontiercs}, EdgeBench~\citep{zhu2026edgebench}, and DiscoGen~\citep{goldie2026discogen} pose open-ended algorithmic problems with graded, unbounded scores.

\xhdr{Test-Time Compute and Population-Based Search}
\emph{Go with the winners} \citep{aldous1994gwtw} keeps a population of
searchers and periodically replaces its weak members with copies of strong
ones. Population Based Training \citep{jaderberg2017pbt} applies the same
exploit-and-perturb cycle to neural-network training, where a worker copies a
better worker's weights and then jitters its hyperparameters. Fork-and-flush
is this scheme over agents: the state copied is a workspace of files rather
than a parameter vector, and the perturbation is discarding the conversation
that produced those files --- the fork must re-derive its intent from the
artifacts, which is what lets it land in a different basin. Successive
halving and racing algorithms \citep{karnin2013almost} likewise spend rounds
of budget on a shrinking survivor set, which is our selection step without
the shared trunk; Go-Explore \citep{ecoffet2021first} archives promising
states and returns to them before exploring, the same
return-to-the-best-workspace move our fan-out makes. \emph{West-of-N} utilizes the best and worst candidates~\citep{pace2024west} to generate preference pairs for self-improving reward models. In tree search, MCTS
\citep{kocsis2006bandit} grows a tree by expanding the nodes its interim
value estimates favour; Tree of Thoughts \citep{yao2023tree} and Language
Agent Tree Search \citep{zhou2024lats} bring that machinery to LLMs at the
granularity of individual thoughts or actions. Fork-and-flush searches the
same way but coarser: a node is a whole agent episode-chain over a workspace,
expansion is a context-cleared fork, and the value estimate is the task's own
grader --- the granularity at which long-horizon runs actually produce
scores. Finally, sampling-based test-time scaling --- AlphaCode's filtered
mass generation \citep{li2022alphacode}, repeated sampling
\citep{brown2024monkeys,snell2024scaling}, and evolutionary program search
\citep{romera2024funsearch} --- resamples complete solutions from scratch,
where fork-and-flush resamples continuations of the best partial work so far.

\xhdr{Plateaus in Long Reasoning} Related forms of stalled exploration have been observed in long-chain reasoning models. \citet{gema2025inverse} construct tasks on which extending a model's reasoning reduces rather than improves accuracy. \citet{marjanovic2025thoughtology} find that DeepSeek-R1 repeatedly revisits previously explored problem formulations, obstructing further exploration, while \citet{pipis2026loop} study a more extreme form of repetition in which reasoning models fall into textual loops. Closest to our setting, \citet{wen2025parathinker} describe \emph{tunnel vision}: early mistakes commit a model to a suboptimal reasoning path, and recovery becomes less likely as the erroneous prefix grows.

Our work complements these studies by examining the analogous phenomenon in long-running autoresearch agents, which repeatedly modify executable artifacts and receive external feedback over hours to several days. In standard long-chain reasoning, each new token is conditioned on the preceding reasoning trace, continually exposing the model to its earlier choices. Our agents, however, periodically compact their conversational histories, replacing the full trace with a summary of the progress so far. One might expect this compaction to weaken dependence on the earlier reasoning path and reopen exploration. We find that it is insufficient: trajectories still plateau at substantially different levels and remain localized in idea basins. \emph{Fork-and-flush} goes further by starting parallel continuations with fresh chat contexts while preserving the accumulated workspace.

\section{Conclusion}
In this work, we introduced \emph{fork-and-flush}, a periodic intervention designed to help autonomous research agents escape localized regions of the solution space, which we term idea basins. We demonstrated that as agents tackle open-ended problems by repeatedly proposing, evaluating, and refining candidate solutions, independent runs often plateau at substantially different scores—a high-variance bottleneck that persists even with considerable additional compute. To overcome this, \emph{fork-and-flush} forks the agent into parallel trajectories that each inherit a copy of the accumulated workspace but begin with fresh chat contexts. After exploring for a fixed horizon, the agent continues from the highest-scoring branch. Across thirteen long-horizon research and engineering tasks under an equal compute budget, \emph{fork-and-flush} consistently outperforms serial baselines, demonstrating that fork-and-flush provides a practical path for scaling test-time compute in autonomous agents. We present exciting avenues for future work. First, the substantial computational expense and time required to evaluate frontier LLMs on long-horizon agentic workflows inherently limit the number of tasks we could practically include in this study. Second, the branched trajectories currently operate independently without coordination. Future work will investigate inter-agent communication and role specialization across branches, as well as scaling our evaluations to a broader suite of tasks and newly released frontier models.

\subsubsection*{Acknowledgments}
This work was done during the first authors' internships at Microsoft Research.

 \newpage

\bibliography{iclr2027_conference}
\bibliographystyle{iclr2027_conference}

\appendix
\section{Ablation: Full Results}
\label{app:ablation}

Table~\ref{tab:ablation-full} is the per-task form of the ablation summarised in
Section~\ref{sec:exp-ablations}. Each entry is the mean final score of that
method at the task's budget, on the task's own scale, so the four columns
are directly comparable within a row but not across rows.

\begin{table}[t]
\centering
\footnotesize
\setlength{\tabcolsep}{5pt}
\caption{\small Ablation on the impact of forking and flushing: mean final score of each method at the task's budget, over the eight single-run baselines (seven on AHC014, p171 and LWE-rec, six on p150) and each method's five draws (four flush-only cells on LLM-SQL; \emph{flush only}: $K{=}1$ with a flush at each forced round; \emph{fork only}: $K{=}4$ branches that continue the trunk's context). Scores are on each task's own scale, so rows are not comparable with each other; the last two rows min--max normalize every run of a task over all four methods before averaging. Best per row in \textbf{bold}, second best in \emph{italics}. The ablation is run at $6{,}000$ credits on \texttt{Borden-inv}, and every column there is read at that budget.}
\label{tab:ablation-full}
\begin{tabular}{l|cccc}
\toprule
\textbf{task} & \textbf{single-run} & \textbf{flush only} & \textbf{fork only} & \textbf{fork-and-flush} \\
\midrule
\texttt{polyomino} & $88.56$ & $85.91$ & $\mathit{90.74}$ & $\mathbf{94.41}$ \\
\texttt{job-shop} & $9.66$ & $10.00$ & $\mathbf{10.43}$ & $\mathit{10.33}$ \\
\texttt{AHC014} & $23.71$ & $23.04$ & $\mathit{24.59}$ & $\mathbf{57.46}$ \\
\texttt{p150} & $123.47$ & $\mathbf{160.80}$ & $128.57$ & $\mathit{129.74}$ \\
\texttt{p171} & $84.26$ & $79.73$ & $\mathit{86.68}$ & $\mathbf{92.79}$ \\
\midrule
\texttt{sphere-pack} & $43.87$ & $43.74$ & $\mathit{44.36}$ & $\mathbf{44.94}$ \\
\texttt{relay-lay} & $72.61$ & $\mathit{72.80}$ & $72.79$ & $\mathbf{72.81}$ \\
\texttt{LWE-rec} & $\mathbf{155.3}$ & $\mathit{148.4}$ & $147.0$ & $144.6$ \\
\texttt{LLM-SQL} & $26.90$ & $\mathit{27.01}$ & $26.93$ & $\mathbf{27.04}$ \\
\texttt{order-add} & $24.60$ & $\mathit{40.73}$ & $30.09$ & $\mathbf{49.52}$ \\
\texttt{Borden-inv} & $59.92$ & $65.85$ & $\mathit{78.02}$ & $\mathbf{80.24}$ \\
\midrule
\emph{normalized avg.\ (5 original tasks)} & $0.37$ & $0.41$ & $\mathit{0.47}$ & $\mathbf{0.63}$ \\
\emph{normalized avg.\ (all 11 tasks)} & $0.45$ & $0.55$ & $\mathit{0.57}$ & $\mathbf{0.71}$ \\
\bottomrule
\end{tabular}
\end{table}

\section{Task descriptions}
\label{app:tasks}
\label{sec:appendix-tasks}

Table~\ref{tab:tasks} lists every task in the study with its source, the
problem the agent optimizes, the score the judge reports, and the credit
budget. Sources are Frontier-CS (FCS; the algorithmic track is cited by
problem number, plus its 2.0 and research tracks), EdgeBench, Anthropic's
public performance-engineering take-home, or a custom task written for this
study. Every task has eight independent single-run baselines
and five fork-and-flush seeds (four on tinyCIFAR). A credit
meters model tokens, not wall-clock.

\begin{table}[h]
\centering
\footnotesize
\setlength{\tabcolsep}{4pt}
\renewcommand{\arraystretch}{1.15}
\caption{\small The thirteen tasks. ``Credits'' is the per-run budget, set per task to where single runs plateau. On \texttt{tinyCIFAR} it is the length of the shortest single-run baseline. ``Hours'' is the median time a single-run baseline takes to reach the budget, counting only time the agent is working and excluding pauses between resumed legs. Most of it is the agent's own solvers and training jobs, which consume no credits; on LWE-rec and order-add single episodes of this kind run for up to two days.}
\label{tab:tasks}
\begin{tabular}{l p{1.8cm} p{4.0cm} p{2.7cm} r r}
\toprule
\textbf{task} & \textbf{source} & \textbf{problem} & \textbf{score} & \textbf{credits} & \textbf{hours} \\
\midrule
\texttt{polyomino}   & FCS algo 0   & Pack a given set of polyominoes into a bounding rectangle of minimal area (C++ heuristic). & Ratio to a reference solution. & 25k & 29 \\
\texttt{job-shop}    & FCS algo 46  & Job-shop scheduling: order operations on machines to minimize the makespan. & Ratio to known bounds. & 10k & 15 \\
\texttt{AHC014}      & FCS algo 159 & Port of AtCoder Heuristic Contest 014. & Ratio to a reference solution. & 10k & 68 \\
\texttt{p150}        & FCS algo 150 & Port of an AtCoder Heuristic Contest problem. & Ratio to a reference solution. & 10k & 70 \\
\texttt{p171}        & FCS algo 171 & Port of an AtCoder Heuristic Contest problem. & Ratio to a reference solution. & 10k & 80 \\
\texttt{kernel}      & Anthropic take-home \citep{anthropic2026takehome} & Anthropic's performance-engineering take-home: optimize a kernel for a simulated VLIW machine; a deterministic simulator counts cycles. & Simulated-cycle score, higher is better. & 100k & 64 \\
\texttt{tinyCIFAR}   & custom       & Train a CIFAR classifier that trades test accuracy against compressed model size (GPU). & Merit in $(0,1)$: an accuracy gate times a log-size term, read on a sealed test split. & 31k & 45 \\
\midrule
\texttt{sphere-pack} & FCS algo 48  & Pack equal spheres in a cube (continuous geometry). & Normalized between a grid baseline and the Kepler bound. & 6k & 20 \\
\texttt{relay-lay}   & FCS algo 307 & Mobile relay layout: $K$-hub facility location with up to $2{\times}10^{5}$ clients. & $10^{6}\,C_{\mathrm{base}}/(C_{\mathrm{base}}+C)$, unclipped above the baseline. & 6k & 6 \\
\texttt{LWE-rec}     & FCS 2.0      & Cryptanalysis: recover the secret of 200 structured-LWE instances drawn from ten families of graded difficulty. & Number of instances solved. & 6k & 83 \\
\texttt{LLM-SQL}     & FCS research & Reorder the rows and columns of an LLM-over-SQL workload to maximize the prefix KV-cache hit rate (deterministic objective). & $(\mathrm{hit}-\mathrm{base})/(1-\mathrm{base})$; 100 is a perfect hit rate. & 6k & 23 \\
\texttt{order-add}   & EdgeBench (optimization) & Black-box permutation optimization over $m{=}1000$ elements against a cost oracle (statistical experimental design); no statement to pattern-match. & Official judge score from the rescaled cost, at most 100. & 6k & 107 \\
\texttt{Borden-inv}  & EdgeBench (scientific) & Scientific inverse problem: recover a 3D groundwater contaminant source from well readings; the agent writes both the forward model and the inverter. & Tier-gated judge score. & 10k & 5 \\
\bottomrule
\end{tabular}
\end{table}

\section{Variance Analysis Plots}
\label{app:variance}

\begin{figure}[p]\centering
\includegraphics[width=0.9\linewidth]{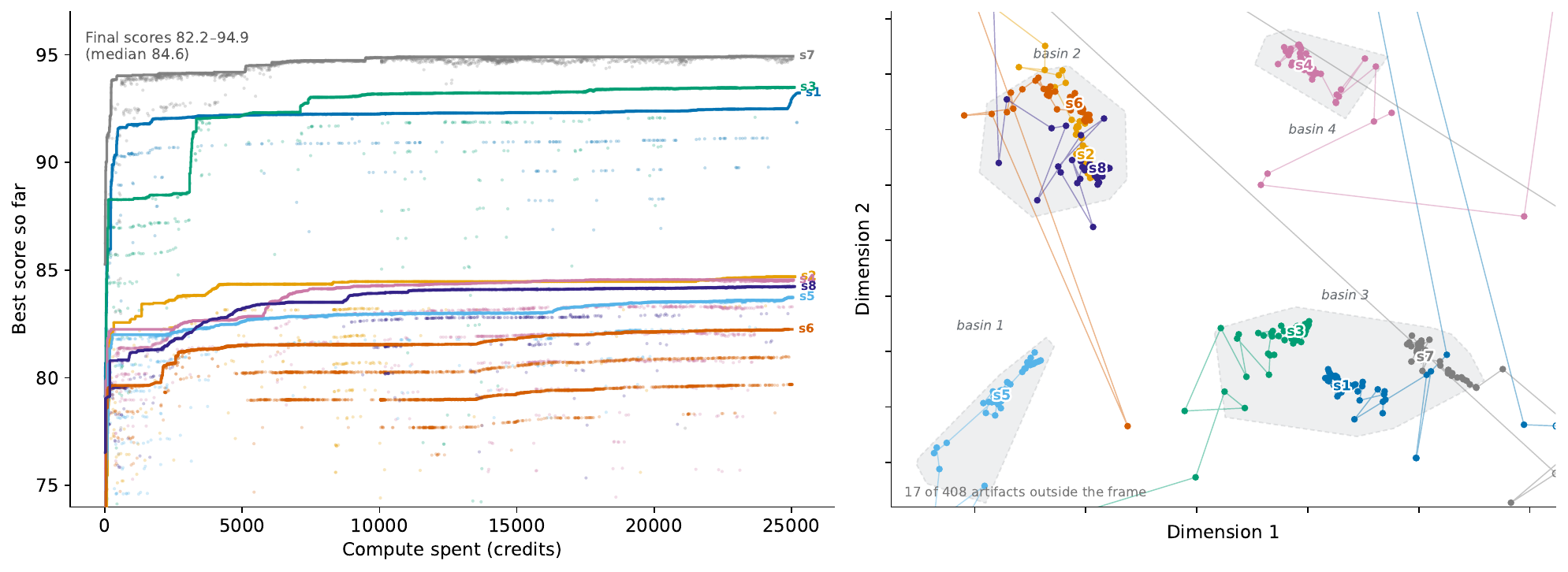}\\[3pt]
\includegraphics[width=0.9\linewidth]{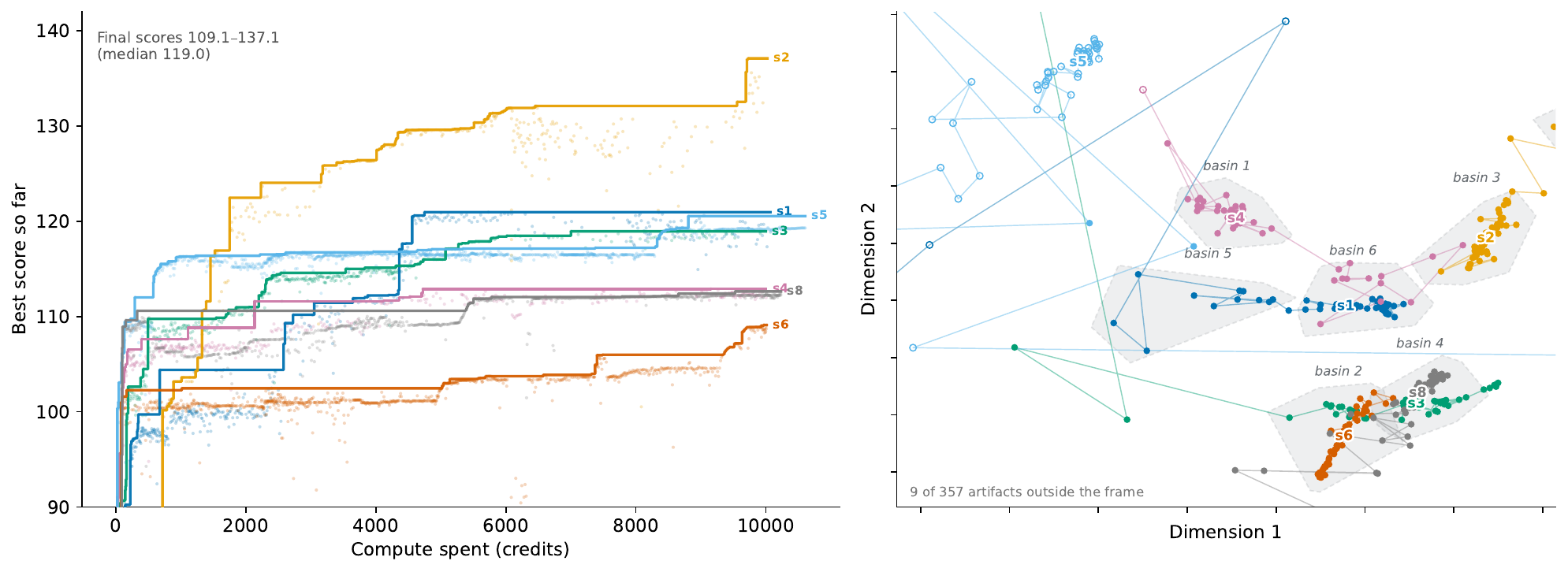}\\[3pt]
\includegraphics[width=0.9\linewidth]{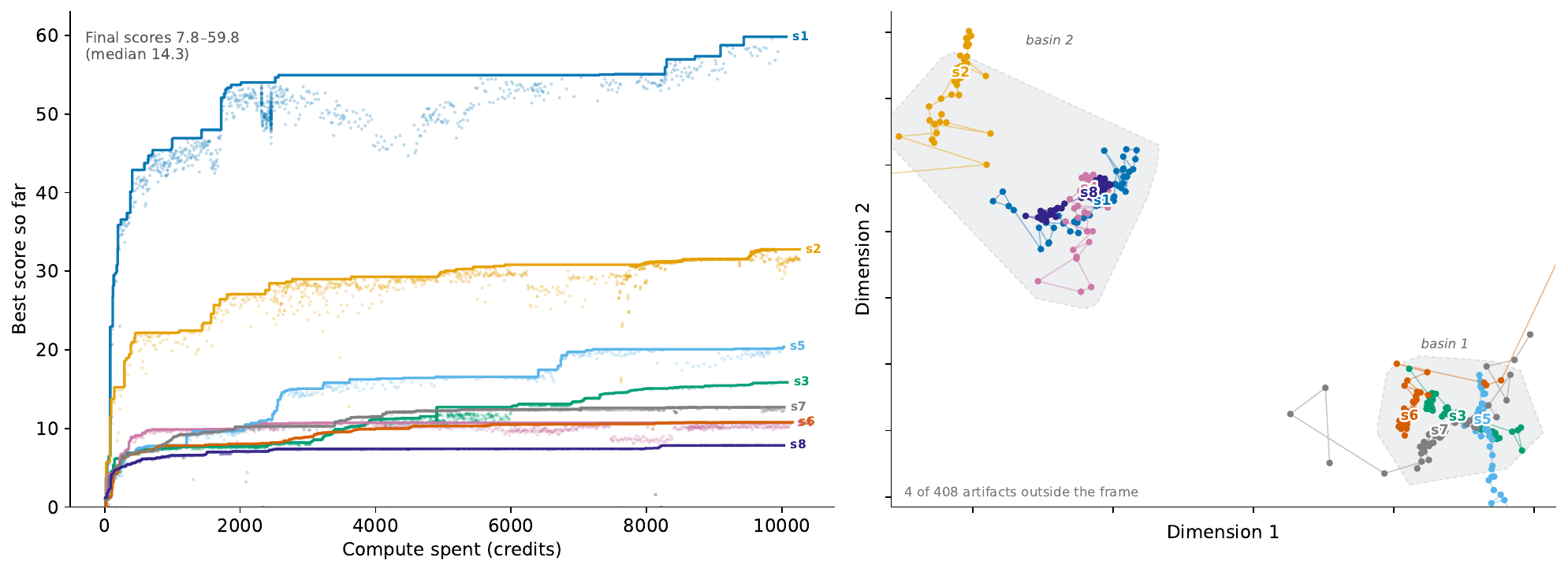}\\[3pt]
\includegraphics[width=0.9\linewidth]{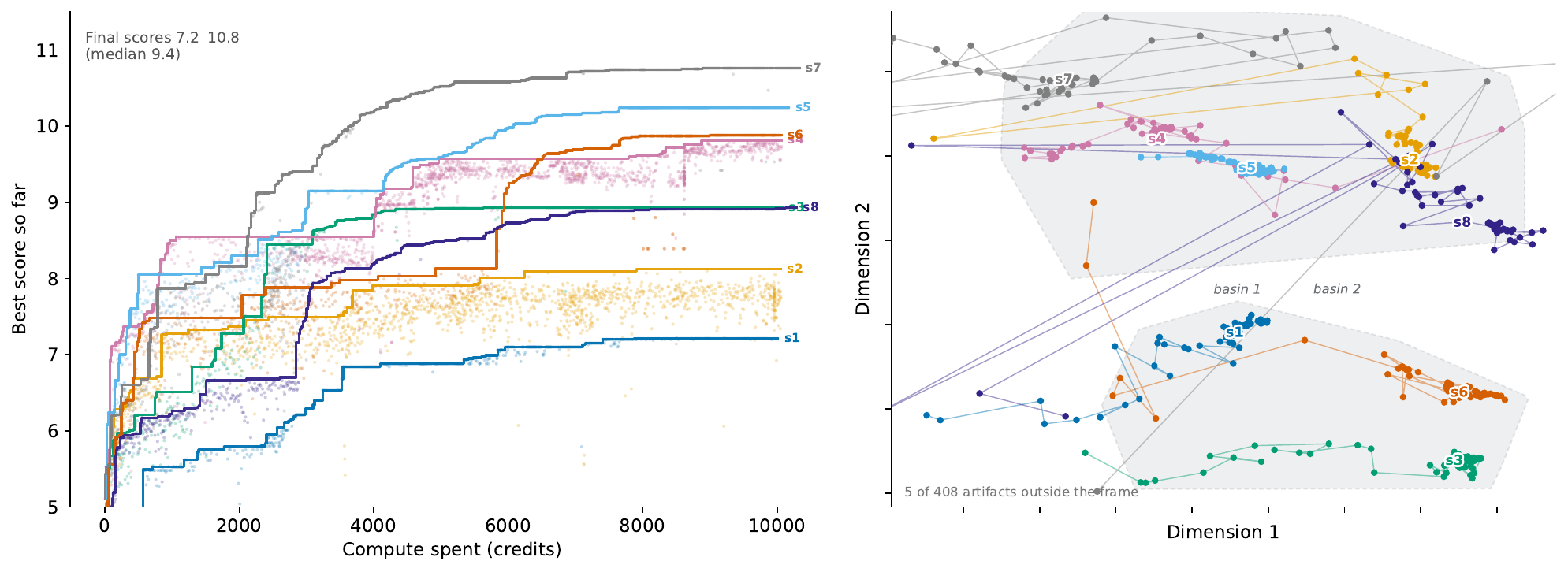}
\caption{\small Independent single runs across four embedded tasks: \texttt{polyomino}, \texttt{p150}, \texttt{AHC014}, and \texttt{job-shop}. \textbf{Left:} Best score achieved versus compute spent. \textbf{Right:} Run artifacts embedded by judged functional similarity, with $k$-means basins shaded.}
\label{fig:baseline-traj-all}
\end{figure}

Figure~\ref{fig:baseline-traj-all} illustrates eight independent single-run trajectories across four embedded tasks. On high-variance problems such as \texttt{polyomino} (final scores $82.2$--$94.9$), \texttt{p150} ($109.1$--$137.1$), and \texttt{AHC014} ($7.8$--$59.8$), the runs clearly separate into distinct basins that track their final performance. For instance, on \texttt{AHC014}, the two highest-scoring runs share a single basin while all remaining runs fall into another. In contrast, on a low-variance task like \texttt{job-shop} ($7.2$--$10.8$), the geometric structure is weak: the two basins overlap heavily, and runs freely move between them. Because this basin structure is poorly defined, we draw no basin-based claims from \texttt{job-shop}.

\section{Schedule Design}
\label{app:schedule_design}

\xhdr{Fixed Schedule vs. Plateau Detection}
The natural trigger for a fork-and-flush round is a performance plateau. However, early development runs using a stall-clock trigger rarely executed the intervention: across five tasks, $15$ of $25$ plateau-triggered runs exhausted their budget before the stall criteria were met. Dynamic triggers proved practically difficult to tune for three reasons. First, credit costs per step vary widely (e.g., $6.1\times$ variance across \texttt{tinyCIFAR} runs), meaning a fixed credit threshold detects a stall prematurely in some runs and entirely misses it in others. Second, tasks often exhibit micro-improvements that restart the stall clock. This allows runs creeping upward within a narrow local basin to evade detection---the exact failure mode fork-and-flush is designed to fix. Third, requiring a sufficiently long stall window (e.g., $20\%$ of the budget) to ignore noise pushes the first intervention too late in the trajectory. A fixed schedule resolves all three issues by guaranteeing the intervention fires, standardizing the branch spend, and removing task-dependent tolerance tuning.

\xhdr{Sensitivity to Cadence}
While we did not systematically sweep all schedule parameters, our development runs span three configurations (Table~\ref{tab:cadence}): the plateau trigger, a single forced round at $50\%$ of the budget, and our main setting of two forced rounds at $33\%$ and $66\%$. Both fixed schedules outperform the single-run baseline on all six tested tasks, with neither strictly dominating (the two-round schedule wins on four tasks, the one-round on two). The plateau trigger performed poorly on \texttt{AHC014} and \texttt{p171} primarily because it never fired. Though sample sizes are small, this suggests the performance gains are robust to the exact cadence of the fixed schedule.

\xhdr{Sensitivity to Branch Count}
We set the branch count to $K{=}4$. Because the branch spend per round is fixed at $K\,b_{\text{explore}}$, increasing $K$ requires either drastically reducing the exploration budget per branch or running fewer rounds entirely. We found that $K{=}4$ strikes a practical balance, allocating roughly $6\%$ of the total budget to each branch---enough depth to meaningfully develop a new idea---while preserving enough global budget for multiple rounds of intervention. The minimal case of $K{=}1$ corresponds to the flush-only ablation discussed in Section~\ref{sec:exp-ablations}.

\begin{table}[h]
\centering
\footnotesize
\caption{\small Fork-and-flush under three schedules, development runs ($K{=}4$). We report the mean final score at the task's budget, with the number of runs in parentheses. The single-run column is the mean of the eight baselines from the same period.}
\label{tab:cadence}
\begin{tabular}{l|r|rrr}
\toprule
\textbf{task} & \textbf{single-run} & \textbf{plateau trigger} & \textbf{one round at 50\%} & \textbf{rounds at 33/66\%} \\
\midrule
\texttt{polyomino} & $87.54$ & $93.84$ (1) & $\mathbf{94.43}$ (1) & $93.82$ (6) \\
\texttt{job-shop} & $9.23$ & $\mathbf{11.44}$ (1) & $10.24$ (2) & $10.99$ (2) \\
\texttt{AHC014} & $21.32$ & $21.30$ (3) & $26.98$ (4) & $\mathbf{40.97}$ (5) \\
\texttt{p150} & $118.24$ & $126.72$ (4) & $\mathbf{145.13}$ (2) & $135.52$ (7) \\
\texttt{p171} & $83.70$ & $80.16$ (4) & $85.83$ (5) & $\mathbf{89.11}$ (11) \\
\texttt{kernel} & $120.36$ & -- & $132.11$ (5) & $\mathbf{136.77}$ (2) \\
\bottomrule
\end{tabular}
\end{table}

\section{Robustness of the Embedding Space}
\label{app:soe}

\newcommand{\rowlab}[1]{\rotatebox{90}{\parbox{0.3966\linewidth}{\centering\small\textbf{#1}}}}
\begin{figure}[!ht]\centering
\rowlab{t-STE}\hfill\includegraphics[width=0.95\linewidth]{figs/fig_basins_jump_tste.pdf}\\[4pt]
\rowlab{SOE}\hfill\includegraphics[width=0.95\linewidth]{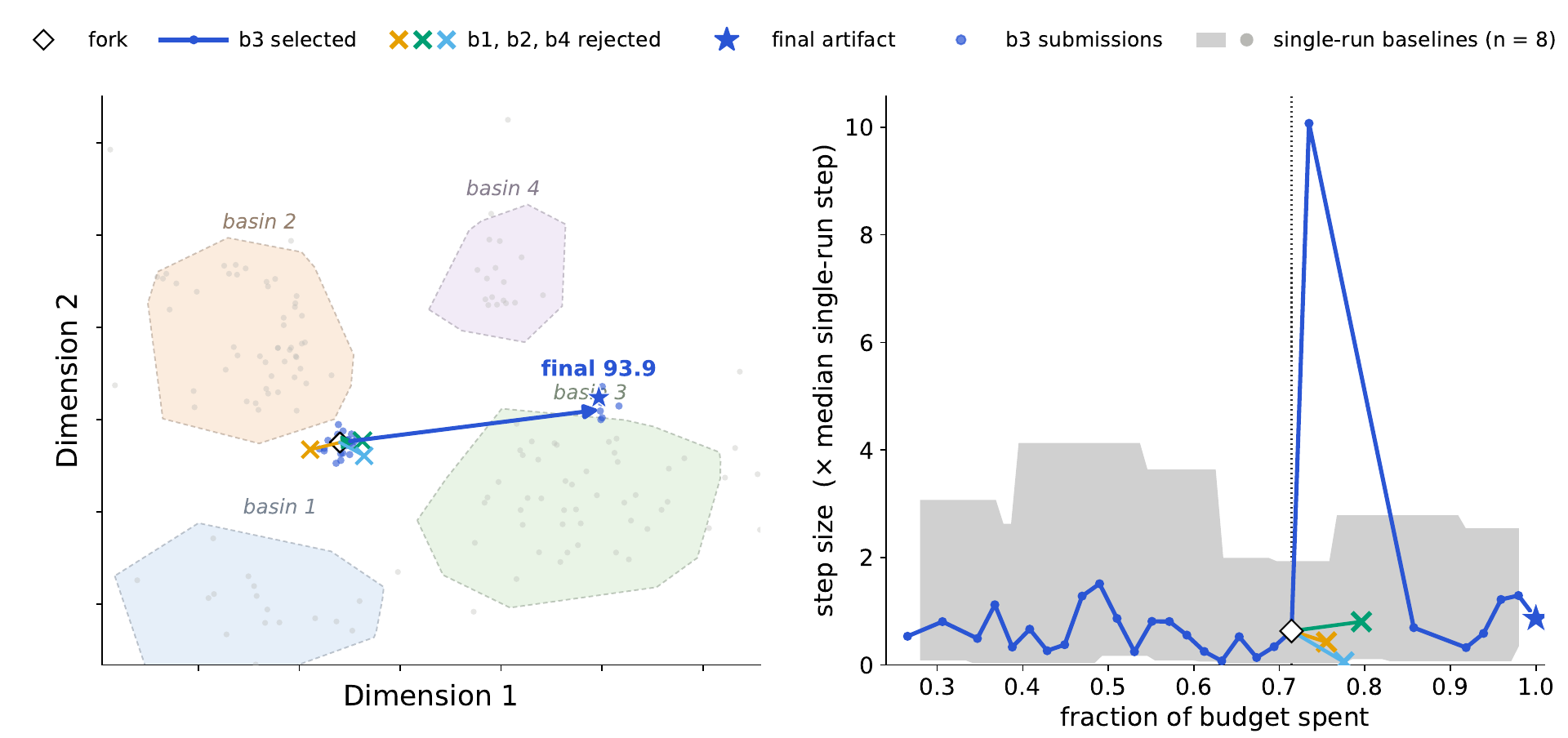}
\caption{\small Figure~\ref{fig:basin-gallery} under two embeddings of the same judged triplets. \textbf{Top:} t-STE, as in Figure~\ref{fig:basin-gallery}. \textbf{Bottom:} SOE, aligned to the same frame, with basins recomputed on its coordinates; $99\%$ of single-run artifacts keep their basin. In both, the rejected branches ($\times$) stay in the branch point's basin and within about one median single-run step of it, while the selected branch lands in a different basin: $43\times$ the median single-run step under t-STE and $10\times$ under SOE, in each case more than twice the largest step any single run takes.}
\label{fig:basin-soe}
\end{figure}

In Section~\ref{sec:analysis-geometric}, we visualized idea basins using t-Distributed Stochastic Triplet Embedding (t-STE) \citep{maaten2012tste}. To verify the robustness of our geometric findings, we refit the exact same LLM-judged triplets using Soft Ordinal Embedding (SOE) \citep{terada2014soe}. Figure~\ref{fig:basin-soe} presents a direct side-by-side comparison of the two embeddings for the \texttt{polyomino} task. The top row reproduces the t-STE map and step distances from Figure~\ref{fig:basin-gallery}, while the bottom row shows the SOE refit aligned to the same frame.  Crucially, the qualitative behavior of the \emph{fork-and-flush} intervention remains identical under the SOE geometry. The rejected branches ($\times$) remain trapped within the local basin of the fork point, moving less than one median single-run step. Conversely, the selected branch successfully jumps to a different idea basin. While the absolute numerical magnitude of the jump scales down under SOE (from $43\times$ the median single-run step under t-STE to $10\times$ under SOE), it remains definitively significant: the jump is still more than twice the largest step any single-run baseline ever takes ($4.1$ units). This confirms that the ability of \emph{fork-and-flush} to escape localized plateaus is a genuine property of the search strategy that is robust to the choice of embedding algorithm.

\section{Artifact Embedding Methodology}
\label{app:mapping}

Each idea-space map is produced in four stages:

\xhdr{(i) Subsampling}
From every trajectory we retain the exact graded submission artifacts, deduplicate identical files by content hash, and select a fixed number of evenly spaced snapshots per run in order of first occurrence ($51$ per run on every mapped task).

\xhdr{(ii) Specification}
An LLM rewrites each sampled submission into a canonical functional specification under a fixed section schema with an exhaustive parameter table. The rewrite strips coding syntax and stylistic signal---formatting, naming, how the file is organized---while preserving the algorithm, data structures, and every numeric choice. This ensures that a competent engineer who has never seen the file could reimplement it from the specification alone and obtain materially the same behavior and score. Judging specifications rather than raw source keeps two artifacts edited from a common ancestor from being called similar merely because they \emph{read} alike.

\xhdr{(iii) Triplet Comparison}
An LLM judge answers ordinal queries---is anchor $a$'s approach closer to $b$'s or to $c$'s?---over the specifications, mixing within-run triplets with cross-condition anchors balanced across conditions. The judge is the same LLM deployment that wrote the specifications, so a single model accounts for every generated token in the pipeline, and each map is built from $6{,}375$ judgments ($5{,}610$ on \texttt{p150}).

\xhdr{(iv) Embedding}
We fit a two-dimensional t-STE embedding \citep{maaten2012tste} to the judged rankings, keeping the best of eight random restarts by training agreement. One in five judged triplets per anchor is held out and never used in fitting; Table~\ref{tab:embedding-heldout} reports agreement on those held-out judgments.

\begin{table}[h]
\centering
\footnotesize
\caption{\small Held-out validation of the idea-space maps. Agreement is the fraction of LLM triplet judgments the fitted two-dimensional map reproduces; chance is $0.50$. t-STE is the embedding used in the figures; SOE refits the same triplets with soft ordinal embedding.}
\label{tab:embedding-heldout}
\begin{tabular}{l|r|rr|rr}
\toprule
& & \multicolumn{2}{c|}{t-STE} & \multicolumn{2}{c}{SOE} \\
task & judged triplets & train & held out & train & held out \\
\midrule
\texttt{polyomino} & 6{,}375 & 0.90 & 0.87 & 0.90 & 0.85 \\
\texttt{AHC014}    & 6{,}375 & 0.90 & 0.87 & 0.91 & 0.86 \\
\texttt{p171}      & 6{,}375 & 0.88 & 0.85 & 0.88 & 0.84 \\
\texttt{p150}      & 5{,}610 & 0.87 & 0.83 & 0.86 & 0.80 \\
\texttt{job-shop}  & 6{,}375 & 0.86 & 0.81 & 0.85 & 0.79 \\
\midrule
\emph{mean} & & 0.88 & 0.85 & 0.88 & 0.83 \\
\bottomrule
\end{tabular}
\end{table}

\xhdr{Validation} Each map is fit on $80\%$ of the judged triplets and scored
on the remaining $20\%$; agreement (Table~\ref{tab:embedding-heldout}) is the
fraction of held-out judgments the two-dimensional coordinates reproduce, with
chance at $0.50$. The t-STE maps reach $0.81$--$0.87$ (mean $0.85$) with a
train--test gap of $0.03$--$0.05$, so they generalise rather than memorise.
Refitting the same triplets with SOE, whose hinge loss does not inflate the
gaps between clusters, gives $0.79$--$0.86$ (mean $0.83$): a slightly weaker
fit, so t-STE is used for the figures and SOE as the check that the
displacements are not artefacts of its
heavy-tailed kernel. Agreement is lowest on job-shop, the task with the least
outcome variance and the weakest basin structure. Run membership alone
(``artifacts from the same run are closer'') reproduces only $0.52$--$0.60$ of
held-out judgments, so the maps capture more than which run an artifact came
from.

\section{Fork-and-flush trajectories}
\label{app:wall}

\begin{figure}[p]
\centering
\includegraphics[width=\linewidth]{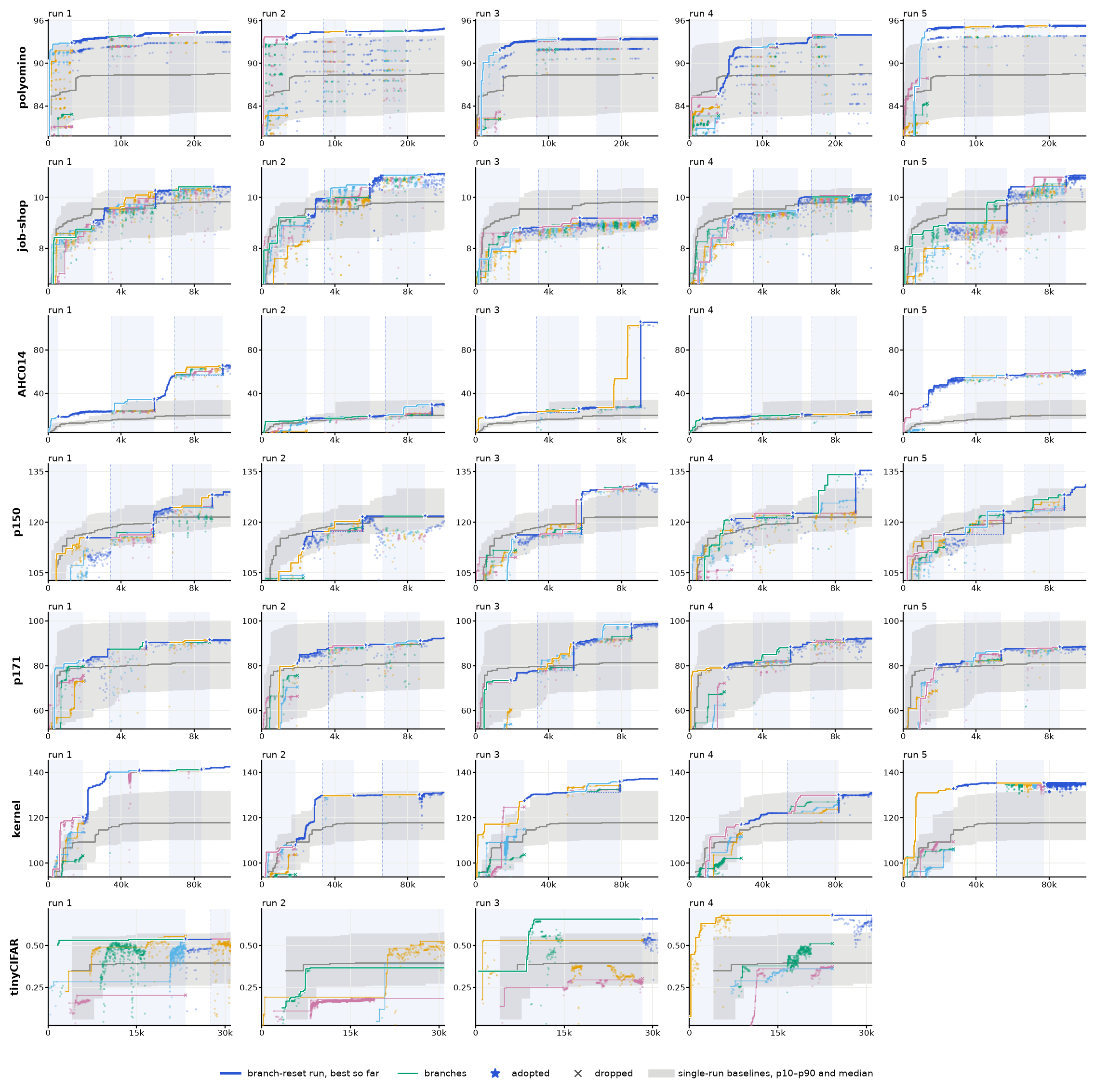}
\caption{Fork-and-flush trajectories against the single-run baselines of each task.}
\label{fig:wall}
\end{figure}

Figure~\ref{fig:wall} shows the trajectories of the fork-and-flush runs in Table~\ref{tab:main-results}, one row per task. Each panel plots score against credits up to the task's budget, the point at which the table reads it. The grey band is the $p_{10}$--$p_{90}$ range of the same single-run baselines the table uses, and the grey line is their median. The blue line is the run's best score so far. Each branch is drawn in its own colour from the score it inherited at the fork, and every graded submission is a dot. A star marks the branch adopted when a round closes, a cross each branch that was dropped, and the shaded columns span the branch rounds. On \texttt{kernel} and \texttt{tinyCIFAR}, some runs use a single forced round at $50\%$ of the budget (Appendix~\ref{app:schedule_design}).

\section{Trajectories Across Idea Basins}
\label{app:jumps}
To further illustrate how the intervention navigates the solution space, Figure~\ref{fig:gallery} visualizes nine distinct \emph{fork-and-flush} trajectories mapped onto their respective task's idea basins. The embeddings reveal three distinct search behaviors. First, during the opening round, initial probes frequently scatter across multiple basins (e.g., \texttt{p171-fd2}, \texttt{ahc014-f2}), allowing the agent to evaluate fundamentally different structural approaches before committing. Second, later in the run, flushes can trigger a definitive jump into a new basin (as seen in \texttt{poly-d4}), enabling the run to successfully escape a local optimum. Finally, even when the flush does not induce a basin jump—and both rejected and selected branches remain localized near the fork (e.g., \texttt{ahc014-t1})—the intervention still drives robust local refinement, consistently driving the trajectory to final scores that surpass the single-run baselines.

\newcommand{\gpanel}[2]{\begin{minipage}[t]{0.32\linewidth}\centering
  #1\\[-2pt]{\scriptsize\texttt{#2}}\end{minipage}}

\begin{figure}[p]\centering
\gpanel{\includegraphics[width=\linewidth]{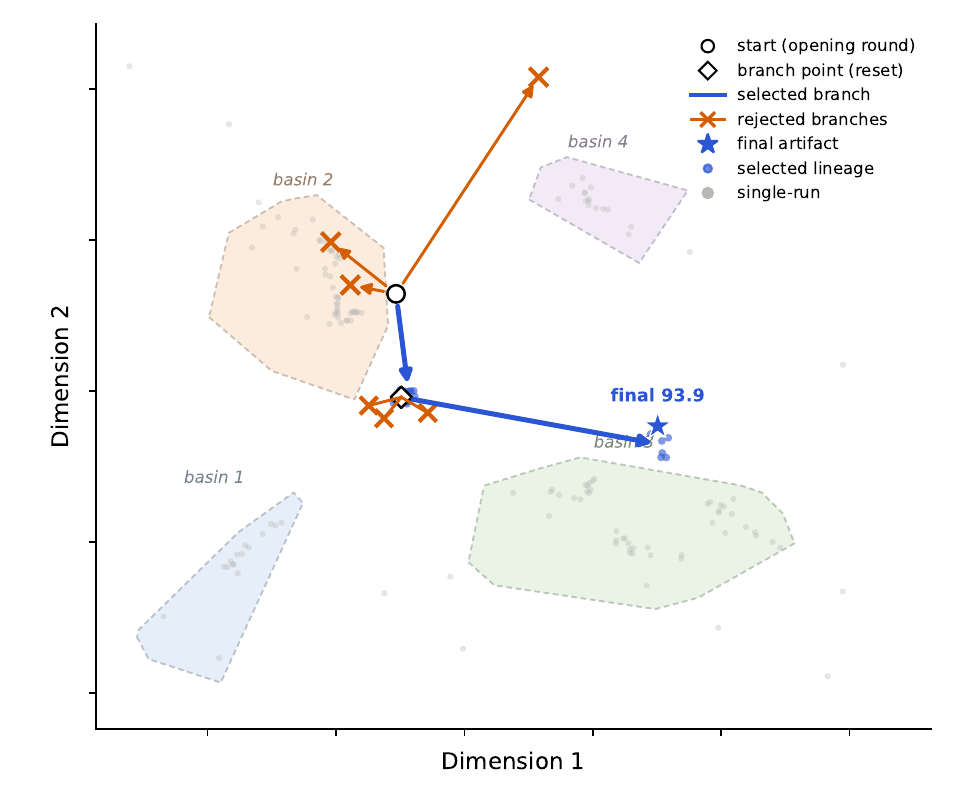}}{poly-d4}\hfill
\gpanel{\includegraphics[width=\linewidth]{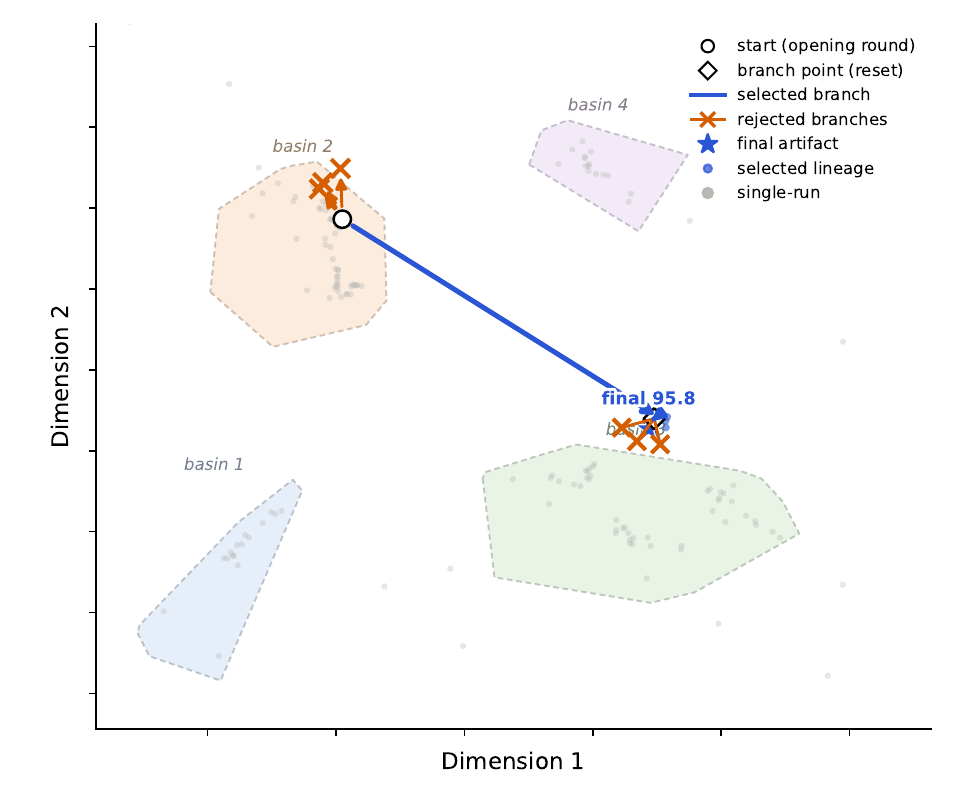}}{poly-a1}\hfill
\gpanel{\includegraphics[width=\linewidth]{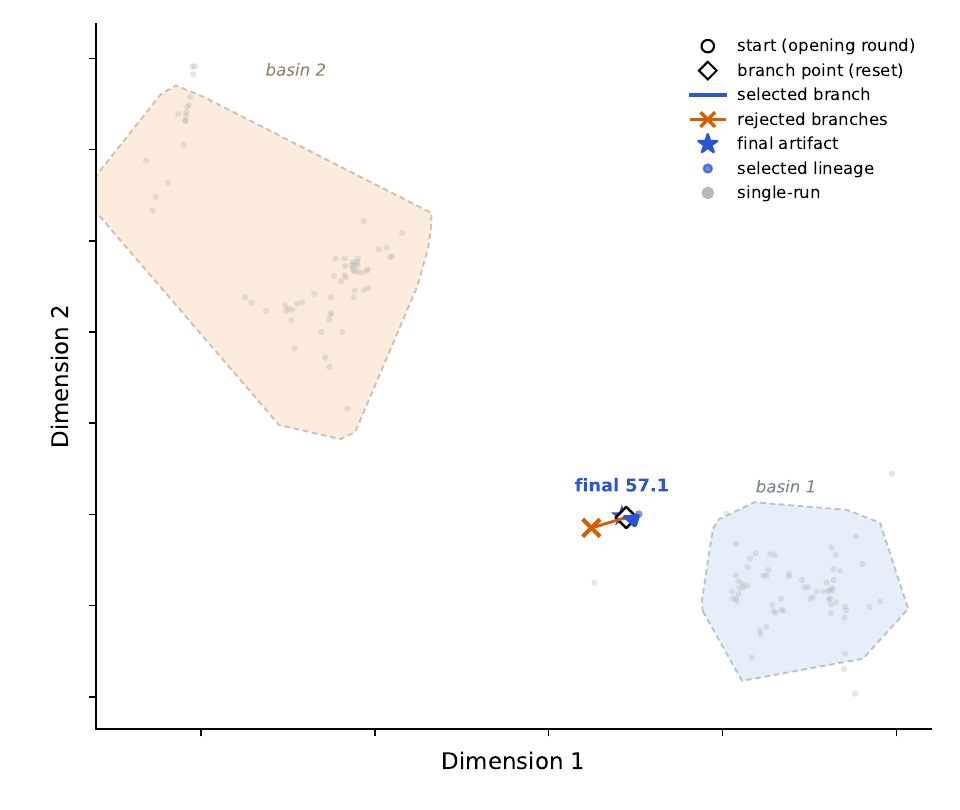}}{ahc014-t4}\\[3pt]
\gpanel{\includegraphics[width=\linewidth]{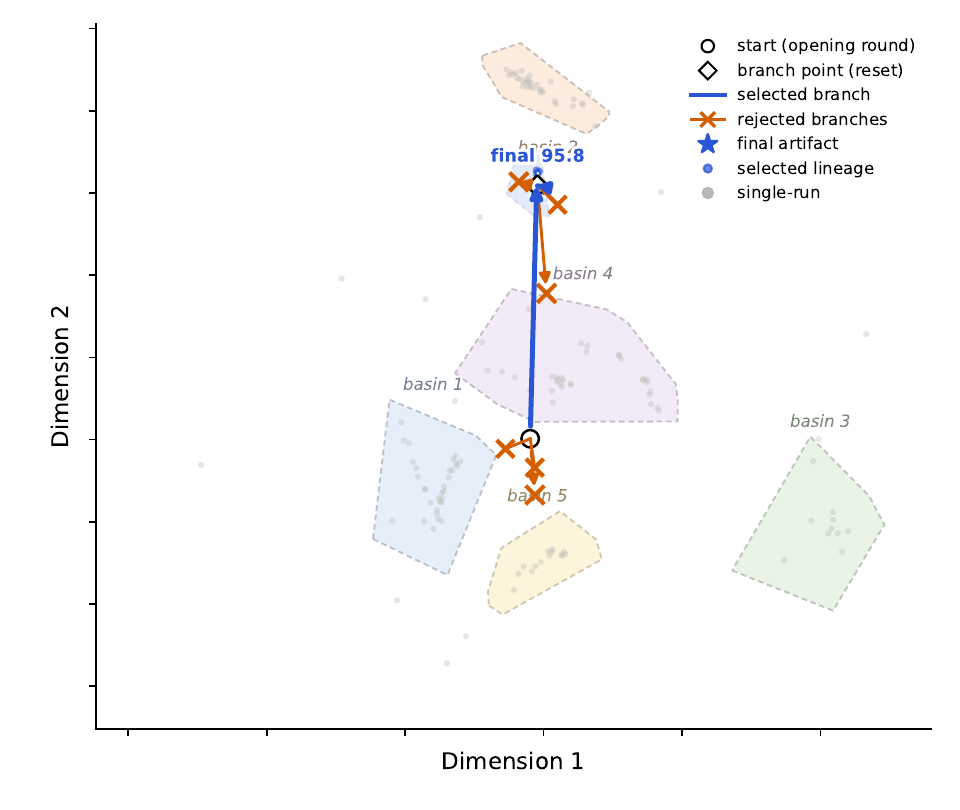}}{p171-fd2}\hfill
\gpanel{\includegraphics[width=\linewidth]{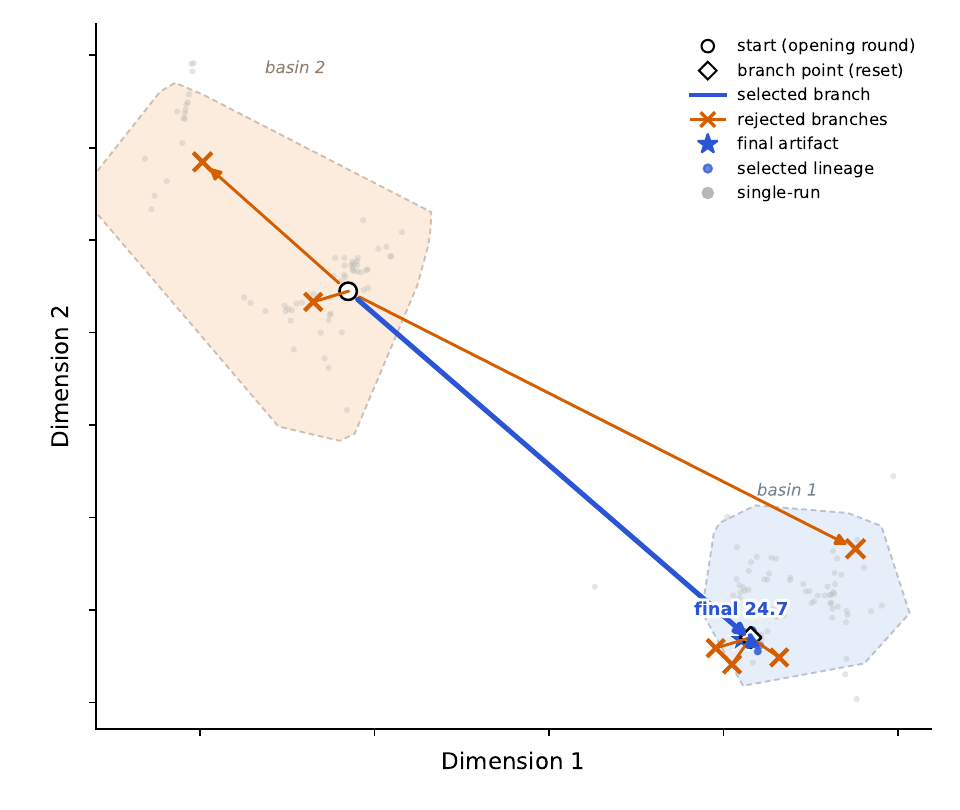}}{ahc014-f2}\hfill
\gpanel{\includegraphics[width=\linewidth]{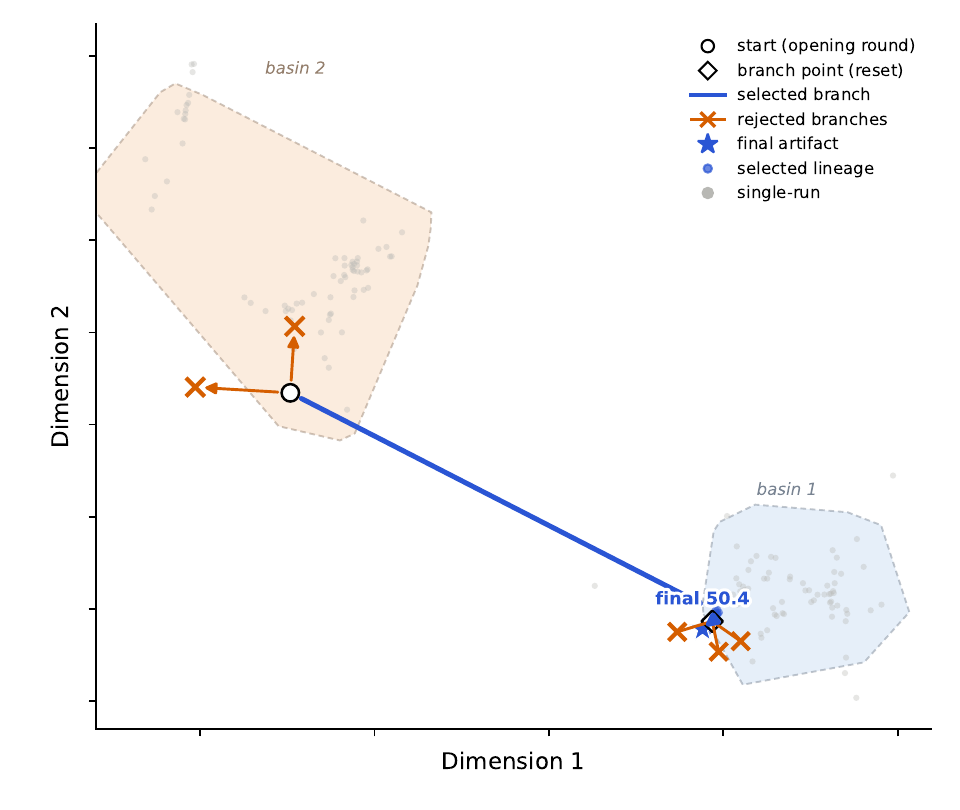}}{ahc014-f4}\\[3pt]
\gpanel{\includegraphics[width=\linewidth]{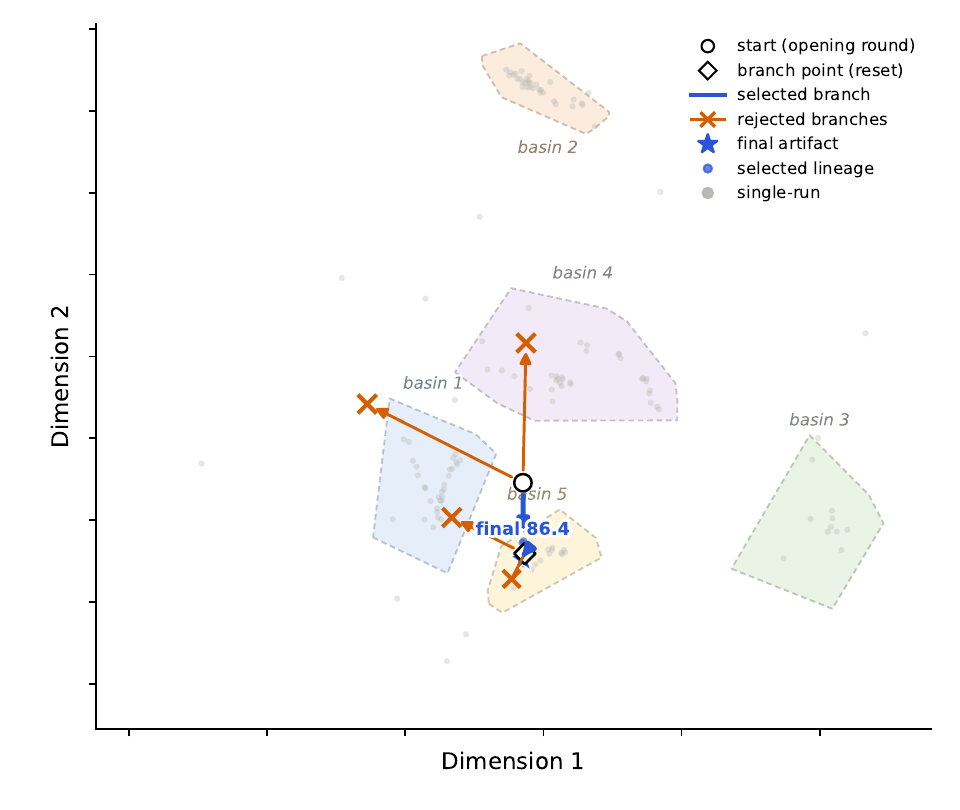}}{p171-fd3}\hfill
\gpanel{\includegraphics[width=\linewidth]{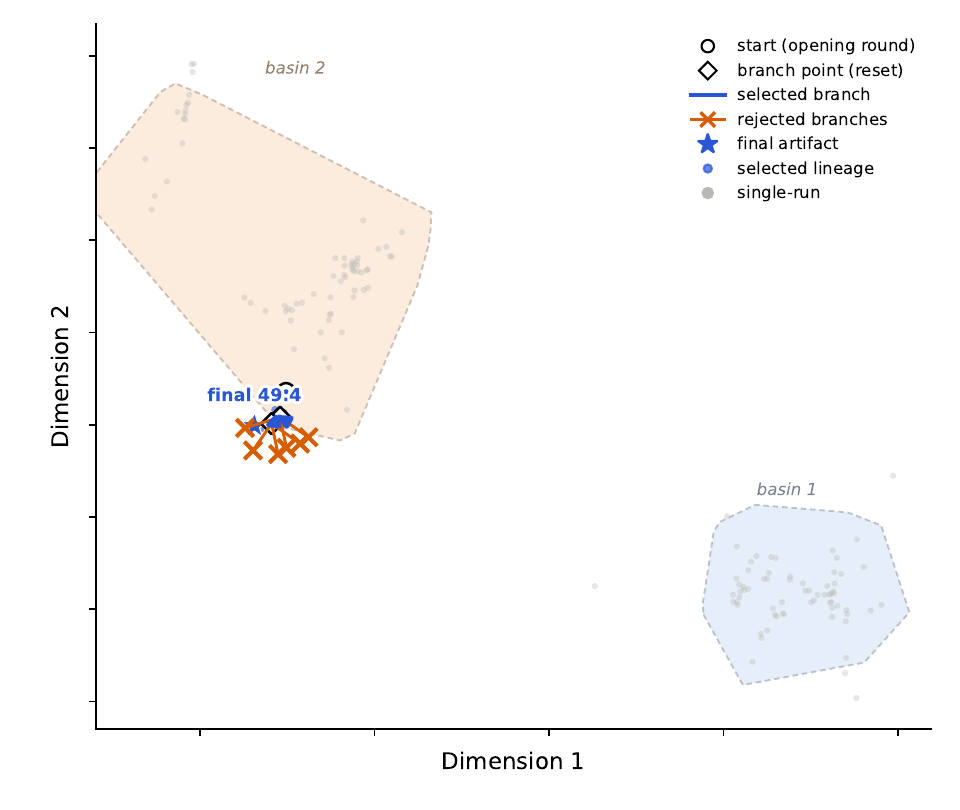}}{ahc014-t1}\hfill
\gpanel{\includegraphics[width=\linewidth]{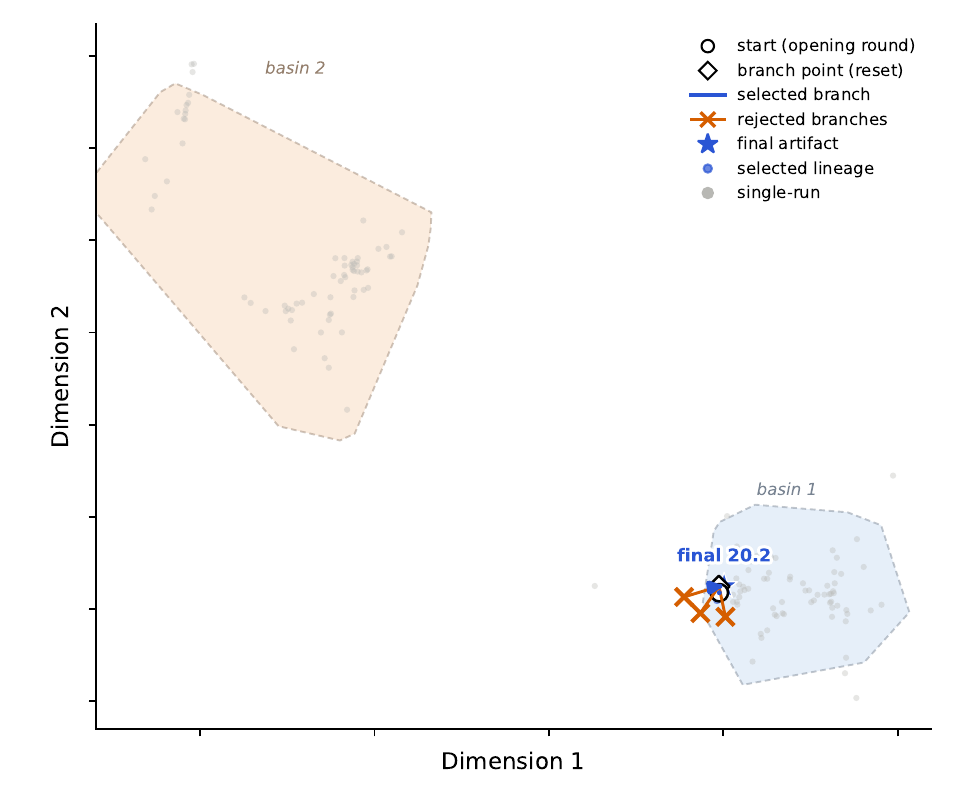}}{ahc014-d3}
\caption{Nine fork-and-flush runs on their tasks' idea maps (task-map basins
shaded). Blue: the selected path ($\circ$ opening round, $\diamond$ fork
with a flush, $\star$ final artifact); vermillion $\times$: rejected branches. }
\label{fig:gallery}
\end{figure}

\clearpage
\section{Prompts}
\label{app:prompts}

\subsection{Agent prompts}

Every message an agent receives is either the task's own packaging (its
statement, its \texttt{AGENT.md} workflow notes and its submission script) or
one of the texts below, which we wrote. No message in any condition states the
credit budget or how much of it remains; agents are never told how much
compute they have. The episode messages are logged verbatim with every run.

\xhdr{Continuation message.} Sent at the start of every episode after the
first, in every condition (single-run, best-of-4 draws, fork-and-flush trunks
and branches), whenever the harness resumes a session:
\begin{promptbox}
Continue --- this task is open-ended and continuously scored; there is no fixed `done'. Keep working in /app. Do not stop.
\end{promptbox}

\xhdr{Branch directive.} Appended to the first message of a directed branch
in a fork-and-flush round. In each round two of the $K$ branches carry it;
the others start from the same workspace with no directive:
\begin{promptbox}
Before continuing, deliberately pick a DIFFERENT approach from the one currently implemented. Do not spend this session incrementally tuning what is already there. Ask whether a fundamentally different algorithm, data structure, or representation would do better, and if it would, implement that instead. It is acceptable for the measured result to get worse before it gets better.
\end{promptbox}
\textbf{The directive does not measurably change which branch wins}: across the $178$ mid-run rounds of the study, a directed branch was selected in $81$ ($46\%$), against the $50\%$ expected if it had no effect (two-sided binomial $p{=}0.26$). We therefore treat the directed and undirected branches as interchangeable.

\xhdr{Protected-incumbent notice.} Appended to the first message of any
branch, or any resumed leg, that inherits a workspace containing a scored
submission. It exists because task scaffolds can overwrite the canonical
solution file, and it is the only place a score is quoted to the agent:
\begin{promptbox}
--- PROTECTED LINEAGE INCUMBENT ---\\
The inherited lineage-best completed submission scored \{score\} and came from episode \{episode\}. Its verified read-only artifact is preserved at \{protected path\} and has been materialized at the canonical path \{canonical path\}. Scaffold, baseline, or bootstrap scripts can overwrite canonical files under /app. Preserve the inherited result or improve it by recovering from the protected artifact; do not silently replace it with the baseline.\\
--- END PROTECTED LINEAGE INCUMBENT ---
\end{promptbox}

\xhdr{Task wrappers.} For the EdgeBench tasks the first message is the text below, followed by the original EdgeBench agent query:
\begin{promptbox}
You are solving a pinned EdgeBench task in \texttt{/app}. The upstream work image's task workspace has been copied to \texttt{/app}. Use \texttt{bash /app/submit.sh} for asynchronous evaluation. The submitted artifact is an exact tar.gz snapshot filtered by the upstream task spec; the hidden judge runs in an isolated disposable copy.
\end{promptbox}
For \texttt{LLM-SQL} it is the text below, followed by the original task statement:
\begin{promptbox}
You are solving the pinned Frontier-CS research task llm\_sql/large. Create \texttt{/app/solution.py}. \texttt{Solution.solve()} may apply the supplied column merges and then change only column order: row order, index, values, dtypes, and exactly-once column membership must remain unchanged. Use \texttt{bash /app/submit.sh} for asynchronous black-box evaluation.
\end{promptbox}
For \texttt{sphere-pack} we replaced the packaging's pass/fail scoring line with the first text below and added the second to its \texttt{AGENT.md}:
\begin{promptbox}
Scoring is continuous: the mean normalized packing ratio across 15 cases; the judge reports the uncapped RatioUnbounded value
\end{promptbox}
\begin{promptbox}
Use fixed random seeds and fixed iteration counts. Do not seed from the clock or terminate search by wall time: exact source artifacts are replayed and must produce the same score.
\end{promptbox}
All other instruction text is the task's own packaging.

\subsection{Idea-map prompts}
\label{app:prompts-map}

The triplet judge's prompt is one template in which a task sentence and a list of the task's typical mechanisms are substituted per task; the \texttt{polyomino} instance is:
\begin{promptlisting}
You are comparing three submitted solutions (A, B, C) to the
same polyomino-packing optimization task, where a submission is a self-contained C++
program that packs sets of polyomino pieces into a rectangular grid to minimize the
enclosed area, scored continuously on how compact its packings are across many problem
instances. Rank the THREE PAIRS -- AB, AC, BC --
by how similar the two members of each pair are in their IMPLEMENTED
OPTIMIZATION STRATEGY, from most similar to least similar.

Judge similarity primarily by computational mechanisms: piece-orientation and symmetry
enumeration, placement search with pruning, width and height selection under area
bounds, portfolio and heuristic construction, backtracking and beam strategies, and
other algorithmic choices.
Ignore naming, comments, formatting, and incidental textual overlap. Do not use
score or chronology; neither is provided. You must produce a strict ordering of
all three pairs even when they are close.

Return STRICT JSON only:
{"order":["AB","AC","BC"],
  "reason":"explain as fully as you need, naming the decisive shared and
            differing mechanisms that place each pair where it is in the order"}
where "order" lists all three pairs from MOST similar to LEAST similar.

=== SUBMISSION A ===
<specification A>

=== SUBMISSION B ===
<specification B>

=== SUBMISSION C ===
<specification C>
\end{promptlisting}
Each task has its own specification prompt with the same eleven-section schema (objective, overall strategy, input handling, representation, the search itself, task-specific selection rules, portfolio and tie-breaking, data structures, control flow and compute budget, an exhaustive parameter table, and techniques not implemented); only the task description and the per-section examples change. The \texttt{polyomino} prompt is:
\begin{promptlisting}
You are writing a canonical FUNCTIONAL SPECIFICATION
of one submitted solution to a polyomino-packing optimization task. The task is
to place a given multiset of polyomino pieces (each an orthogonally-connected
set of unit cells) into a rectangular grid without overlaps, minimizing the
enclosed area (width times height); a submission is a self-contained C++ program
scored continuously on how compact its packings are, aggregated across many
problem instances of different sizes.

The specification has one purpose: a competent engineer who has never seen this
file must be able to reimplement it from your text alone and obtain materially
the same behaviour -- the same search strategy and materially the same packing
quality (area / score). Completeness is the priority. There is no length limit;
do not summarise, do not economise, and do not omit a decision because it seems
minor.

Two solutions that implement the same thing must receive near-identical
specifications even if the files look nothing alike, and two solutions that
implement different things must receive visibly different specifications even
if the files look almost identical. Everything below serves that goal. In this
corpus the differences between solutions are usually in the SEARCH HEURISTICS
(placement order, candidate scoring, tie-breaking, which grid widths are tried,
lookahead depth, portfolio of retried configurations), so those must be captured
in exact detail; a difference there is a difference in the solution.

ABSOLUTELY EXCLUDE -- these are properties of the file, not of the solution:
identifier, function, class, method and variable names; how the code is split
into functions, lambdas, structs or modules; the order definitions appear in;
comments, docstrings, logging and their wording; formatting, style and code
quality; unreachable, unused, vestigial or commented-out code; anything about
the author, the file's history, or how it was developed. Never quote the source.
Refer to every component by its ROLE.

ALWAYS INCLUDE -- these are properties of the solution: every algorithmic
decision, every data-layout decision, and EVERY quantitative choice with its
exact value. Numeric parameters -- candidate widths and how they are chosen,
lookahead / beam widths, area or bounding-box bounds, piece-ordering keys,
tie-break selectors, thresholds keyed on instance size, time or iteration
limits, magic constants in scoring functions -- are FUNCTIONAL CONTENT and must
all be reported with their value and their role. A reimplementation that guessed
these would not reproduce the packing quality.

Emit exactly the following sections, in exactly this order, with these exact
headings, even when a section is empty. Never reorder them to match the file.
If the solution does not do something a section asks about, write
"Not implemented." for that section rather than omitting it -- the absences are
as diagnostic as the presences.

## 1. OBJECTIVE
The computation the program must produce and how it is scored, stated
independently of how (pack the given pieces into a grid without overlap;
minimize enclosed area; scored on compactness across instances).

## 2. OVERALL STRATEGY
The core idea in 3-6 sentences: what this solution does that a naive one does
not (e.g. greedy skyline placement over several candidate widths, exact/backtrack
search on small instances, a portfolio of retried heuristics keeping the best),
and why that produces tighter packings.

## 3. INPUT PARSING AND INSTANCE HANDLING
How the instance is read (number of pieces, per-piece cell lists); whether
multiple instances or sizes are handled and how they are dispatched; any
size-based forking that selects different strategies or parameters for small
versus large instances, with the exact size thresholds.

## 4. PIECE REPRESENTATION AND ORIENTATION ENUMERATION
How each piece is stored; how orientations are generated (rotations, reflections,
how many, and how symmetric duplicates are removed); the per-orientation derived
data computed (bounding box, per-row cell counts, row bitmasks, bottom-contour
cells) and what it is used for.

## 5. PLACEMENT AND PACKING SEARCH
The packing procedure: how the occupied region is represented, how a legal
position for a piece is found (bottom-left / skyline / column-scan / bitmask
collision test), the order pieces are placed, whether placement is greedy,
backtracking, beam or exact, and any lookahead and its depth. Give the exact
scoring/selection rule for choosing among candidate placements. State explicitly
whether the search is DETERMINISTIC or RANDOMIZED: if randomized, report the use
of a seeded RNG, random restarts, shuffled piece/candidate orders, or
sampling-and-keep-best, the number of restarts or samples, and how the best
result is retained -- deterministic versus randomized-best-of-N is a first-order
difference between solutions in this corpus and must never be left implicit.

## 6. WIDTH, HEIGHT AND AREA SELECTION
Which grid widths are tried and how that set is derived (area lower bounds, piece
spans, enumerated ranges); how height results from a packing; how the final area
is chosen among tried widths; any pruning against a current-best area bound.

## 7. PORTFOLIO, HEURISTIC VARIANTS AND TIE-BREAKING
Any set of distinct configurations run and compared (multiple tie-break orders,
orientation orders, candidate orders, root choices, random seeds), how many, how
each is selected, and how the best result across them is retained. Enumerate the
distinct tie-break / ordering rules and what each changes. Distinguish a fixed
enumerated portfolio from a randomized ensemble of restarts.

## 8. DATA STRUCTURES AND BIT-LEVEL REPRESENTATION
Every significant data structure: bitset/word layout of rows, skyline or column
height arrays, candidate lists, caches of best packings, and how they are
addressed and sized.

## 9. CONTROL FLOW AND COMPUTE BUDGET
The end-to-end order of phases that produces the final packing, expressed as a
sequence of steps with their loop bounds and any early exits. Report any
COMPUTE BUDGET that governs how long the search runs: wall-clock time limits,
iteration or restart caps, and any ANYTIME behaviour (returning the best packing
found so far when a deadline is hit). A time-bounded anytime solver and a
fixed-work solver are materially different and this section must say which.

## 10. PARAMETERS
An exhaustive table, one row per numeric or categorical parameter, sorted
alphabetically by role name. Columns: role | value | what it controls | what
would change if it changed. Include size thresholds, lookahead/beam widths,
candidate-width bounds, tie-break selector values, RNG seeds and restart counts,
time or iteration limits, and any scoring constants. Do not skip a parameter
because it looks conventional.

## 11. NOT IMPLEMENTED
Techniques a strong solution to this task might use and that this solution
demonstrably does NOT use (e.g. exact ILP/branch-and-bound, simulated annealing,
per-instance width tuning, rotation-aware nesting, no-fit polygons). Be specific
and honest; this section distinguishes solutions that share a strategy from
solutions that share an ancestor.

If the source below is visibly truncated, specify everything determinable from
what is present and state plainly in the affected section that the remainder
could not be determined. Do not guess at missing code.

Output the specification only, beginning with "## 1. OBJECTIVE", with no
preamble and no closing commentary.

=== SUBMISSION ===
<submission source>
\end{promptlisting}

\end{document}